\documentclass[journal]{IEEEtran}
\usepackage[dvips]{graphicx}

\usepackage{amsfonts,epsfig,cite,array,multirow,graphicx,amsmath,ltablex,tabularx,setspace,arydshln,amssymb,multirow}

\usepackage{subfigure,calc,amsmath,amstext,amsthm,pslatex,multicol}

\usepackage{colortbl,graphics,url}
\usepackage{adjustbox}
\usepackage{array}
\usepackage{changes}
\usepackage{algorithm}
\usepackage{algpseudocode}
\usepackage{booktabs}
\usepackage{bigstrut}

\ifCLASSINFOpdf
\else
\fi
\begin{document}
%
\title{Temporal Unknown Incremental Clustering (TUIC) Model for Analysis of Traffic Surveillance Videos}
%
%
%

\author{
		Kelathodi K. Santhosh, ~\IEEEmembership{Student Member,~IEEE}, Debi P. Dogra,~\IEEEmembership{Member,~IEEE}, and Partha P. Roy
\thanks{
		Kelathodi K. Santhosh and Debi P. Dogra are with the School of  Electrical Sciences, Indian Institute of Technology Bhubaneswar, India 752050.
		Email:\{sk47, dpdogra\}@iitbbs.ac.in
		}
\thanks{
		Partha P. Roy is with the Department of Computer Science and Engineering, Indian Institute of Technology, Roorkee, India.
		Email:proy.fcs@iitr.ac.in
}
}

\maketitle

\begin{abstract}
Optimized scene representation is an important characteristic of a framework for detecting abnormalities on live videos. One of the challenges for detecting abnormalities in live videos is real-time detection of objects in a non-parametric way. Another challenge is to efficiently represent the state of objects temporally across frames. In this paper, a Gibbs sampling based heuristic model referred to as Temporal Unknown Incremental Clustering (TUIC) has been proposed to cluster pixels with motion. Pixel motion is first detected using optical flow and a Bayesian algorithm has been applied to associate pixels belonging to similar cluster in subsequent frames. The algorithm is fast and produces accurate results in $\Theta(kn)$ time, where $k$ is the number of clusters and $n$ the number of pixels. Our experimental validation with publicly available datasets reveals that the proposed framework has good potential to open-up new opportunities for real-time traffic analysis.
\end{abstract}

%
\IEEEpeerreviewmaketitle
\section{Introduction}
\IEEEPARstart{R}{eal} time surveillance is posing a big challenge to the researchers since number of cameras are increasing in leaps and bounds. It is a difficult task to employ a large number of human operators for monitoring huge amount of visual data. Also, it is not possible for human observers to detect all abnormal activities as humans face difficulty to maintain certain level of alertness for a sustained period. Hence automated methods for continuous analysis of surveillance videos have to be introduced. However, while developing such methods, it is a prerequisite to detect abnormal/unusual events on a timely manner such that suitable actions can be taken at the earliest. Moreover, surveillance systems are often equipped with large number of cameras that produce enormous amount of data. Therefore, online storing (while recording) of event(s)-of-interest or useful segments for future processing, can be adopted. This emphasizes the importance of real-time video event detection. 
    
One of the ways to represent video events, is to detect patterns of movement of dynamic objects. In order to detect events in real-time, object(s)-in-motion need(s) to be represented in an analysis framework. Then, a suitable algorithm can be developed to classify the events. In our work, we propose a nonparametric model derived from DPMM and have devised a distance based unsupervised learning scheme to localize moving objects on the run.  Our proposed algorithm has been found to produce real-time performance on publicly available surveillance videos.
       
\subsection{Related Work}
\label{sec:Related_Work} 
There are a few interesting proposals for modeling the object motion that are based on optical flow~\cite{2,3}. These frameworks are  offline and they do not use object trajectories as input signal. However, a few popular methods use trajectory as the base-level information~\cite{4,5,9,13,14,IncrementalDPMM,16,17}. Consequently, trajectory learning and classification are two of the central tasks for any video analytic method. There are studies on supervised approaches such as~\cite{4,5,6,7,8,9,10} that are based on labeled dataset. Unsupervised approaches such as~\cite{11,12,13,14,IncrementalDPMM} use unlabeled dataset to cluster similar trajectories and use clustered data to train models for classification. Tracking~\cite{TLD,KCF,tomasi1991detection,comaniciu2000real} is also an important task to build a complete traffic analysis framework. A recent work proposed in~\cite{24} defines an offline model for tracking using Dirichlet process that is based on variational inference~\cite{26}. An approach often referred to as incremental clustering, has been used in~\cite{IncrementalDPMM} and~\cite{16}. The method works in the absence of complete training data. They have processed the data sequentially. The approach is particularly relevant in surveillance applications since training data may not be available always.

The learned model representing  trajectory patterns can be used for varying purposes irrespective of the underlying method of training. A few of these methods address abnormality detection~\cite{16,17,18},  while others perform classification and abnormality detection together~\cite{4,8}. Trajectory retrieval~\cite{IncrementalDPMM,1} is another possible application. An important property of such a framework is, online classification and abnormality detection with partially observed trajectories. The problem has been addressed in~\cite{4,8,9}. This is important when timely actions are to be taken in response to an observed event.
\begin{figure}[h]
  \centering
  \subfigure[Original frame]{\includegraphics[scale=0.36]{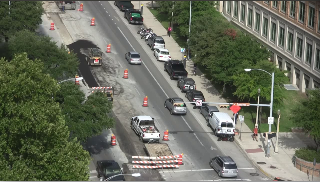}}
  \subfigure[Optical flow magnitude]{\includegraphics[scale=0.36]{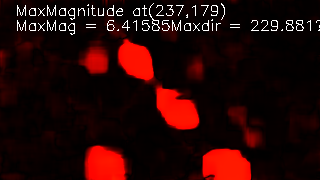}}
  \subfigure[Optical flow direction]{\includegraphics[scale=0.36]{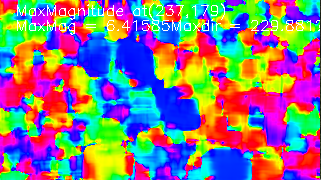}}
  \subfigure[Labeled scene]{\includegraphics[scale=0.36]{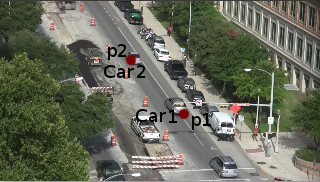}}
  \caption{Optical flow and association of pixels with moving objects.}
  \label{Fig:OpticalFlow}
\end{figure}
\vspace{-.1cm}
 \subsection{Motivation of the Research}
\label{sec:Motivation} 
 An unsupervised, non-parametric, incremental, real-time framework for event detection can be a good choice for surveillance applications as it will reduce the dependency on human operators. However, the problem is challenging due to the complexity of scene interpretation. This is more difficult as everything needs to be built from pixel-level information. The work proposed in~\cite{IncrementalDPMM} and~\cite{16} can be adopted for online learning. However, their model is complex and it requires large number of iterations to cluster pixels to form the objects.

Fig.~\ref{Fig:OpticalFlow} depicts optical flow in a frame taken from the VIRAT dataset~\cite{Dataset}. Fig.~\ref{Fig:OpticalFlow} (a) represents the original frame and Fig.~\ref{Fig:OpticalFlow} (b) and Fig.~\ref{Fig:OpticalFlow} (c)  represent magnitude and direction of flow, respectively. We are motivated by visual clue presented in the figures. It has been observed that the pixels in motion have a distribution similar to a multivariate Gaussian process. Consider two pixels $p_1$ and $p_2$ as marked in Fig.~\ref{Fig:OpticalFlow} (d). Probability of $p_1$  belongs to ``car \#1" is expected to be higher than it belongs to ``car \#2" as $p_1$ is closer to ``car \#1" than ``car \#2". Similarly for $p_2$, probability of this pixel belongs to ``car \#2" is expected to be higher than it belongs to ``car \#1". We have used this distance information for deriving an inference scheme that is fast and logical to be applied on Dirichlet Process Mixture Model (DPMM)~\cite{Rasmussen} based on Dirichlet Process~\cite{ferguson1973}.

Inference process~\cite{25,blei2006} in DPMM takes multiple iterations for clusters to converge. We have introduced a distance function in the inference process of DPMM to expedite the convergence. Distance Dependent Chinese Restaurant Process (DDCRP)~\cite{30} underlines the usage of distance in the inference process for faster convergence. In DPMM, the number of clusters formed on a given data depends on the concentration parameter ($\alpha$) of the model described in~(\ref{equation:DPM1}-\ref{equation:DPM4}). 

\begin{equation}
z_i|\pi  \sim  \mbox{Discrete}(\pi)
\label{equation:DPM1}
\end{equation} 
\begin{equation}
x_i|z_i, \theta_k  \sim  F(\theta_{z_i})
\label{equation:DPM2}
\end{equation} 
\begin{equation}
\pi = (\pi_1, \cdots ,\pi_K)|\alpha  \sim  \mbox{Dirichlet}(\alpha / K, \cdots, \alpha / K)
\label{equation:DPM3}
\end{equation} 
\begin{equation}
\theta_k|H  \sim  H
\label{equation:DPM4}
\end{equation} 

\begin{figure}[h]
  \centering
  \subfigure[Conventional Dirichlet Process Mixture Model (DPMM).]{\includegraphics[scale=0.46]{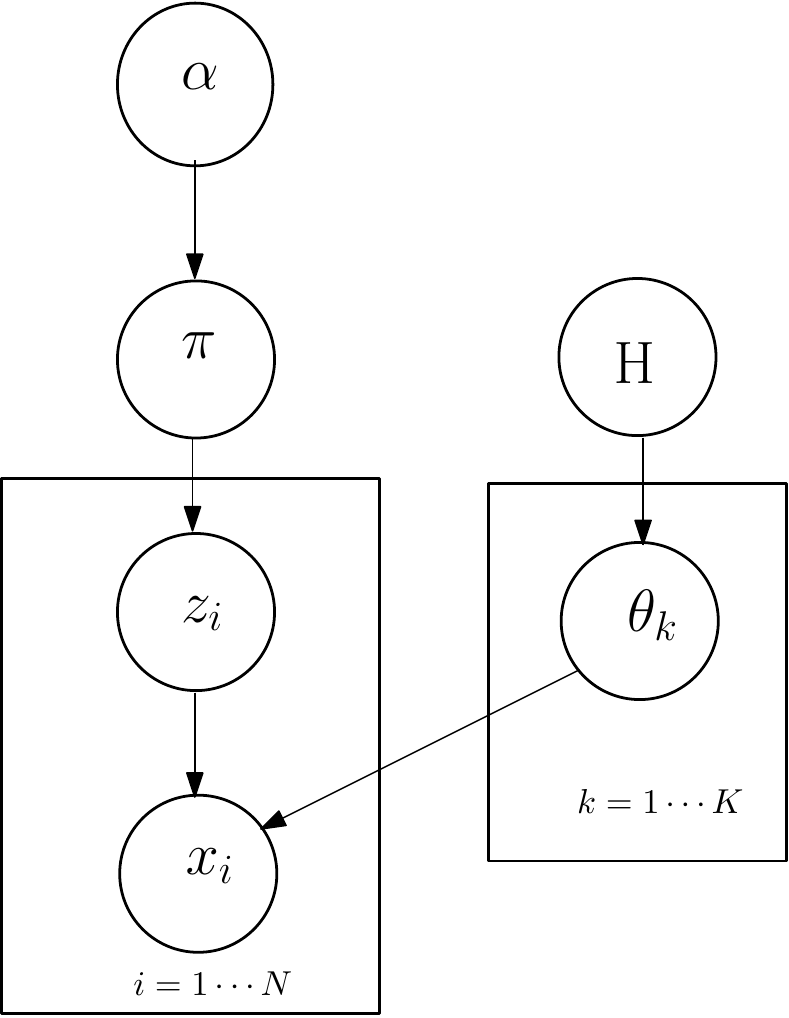}}
  \subfigure[Proposed Object Model]{\includegraphics[scale=0.65]{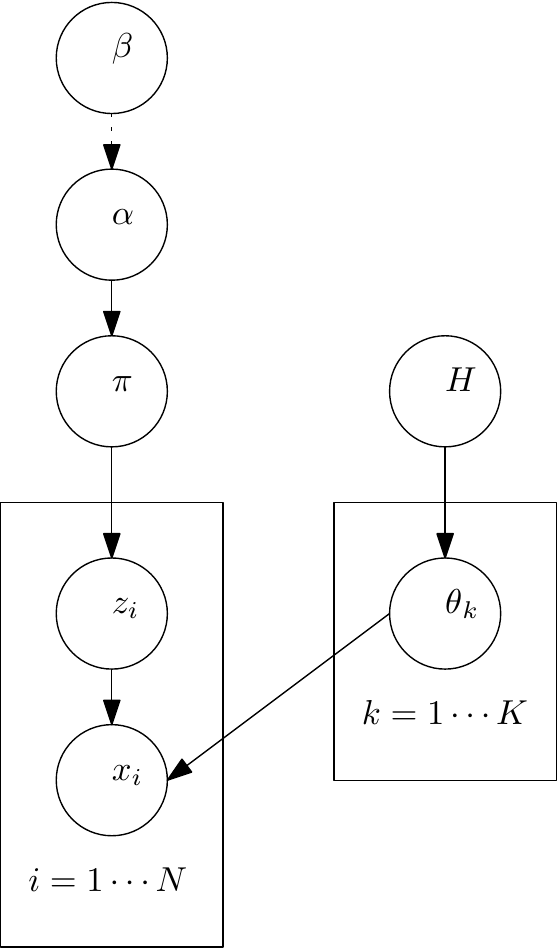}}
  \subfigure[Proposed Temporal Unknown Incremental Clustering (TUIC) Model]{\includegraphics[scale=0.6]{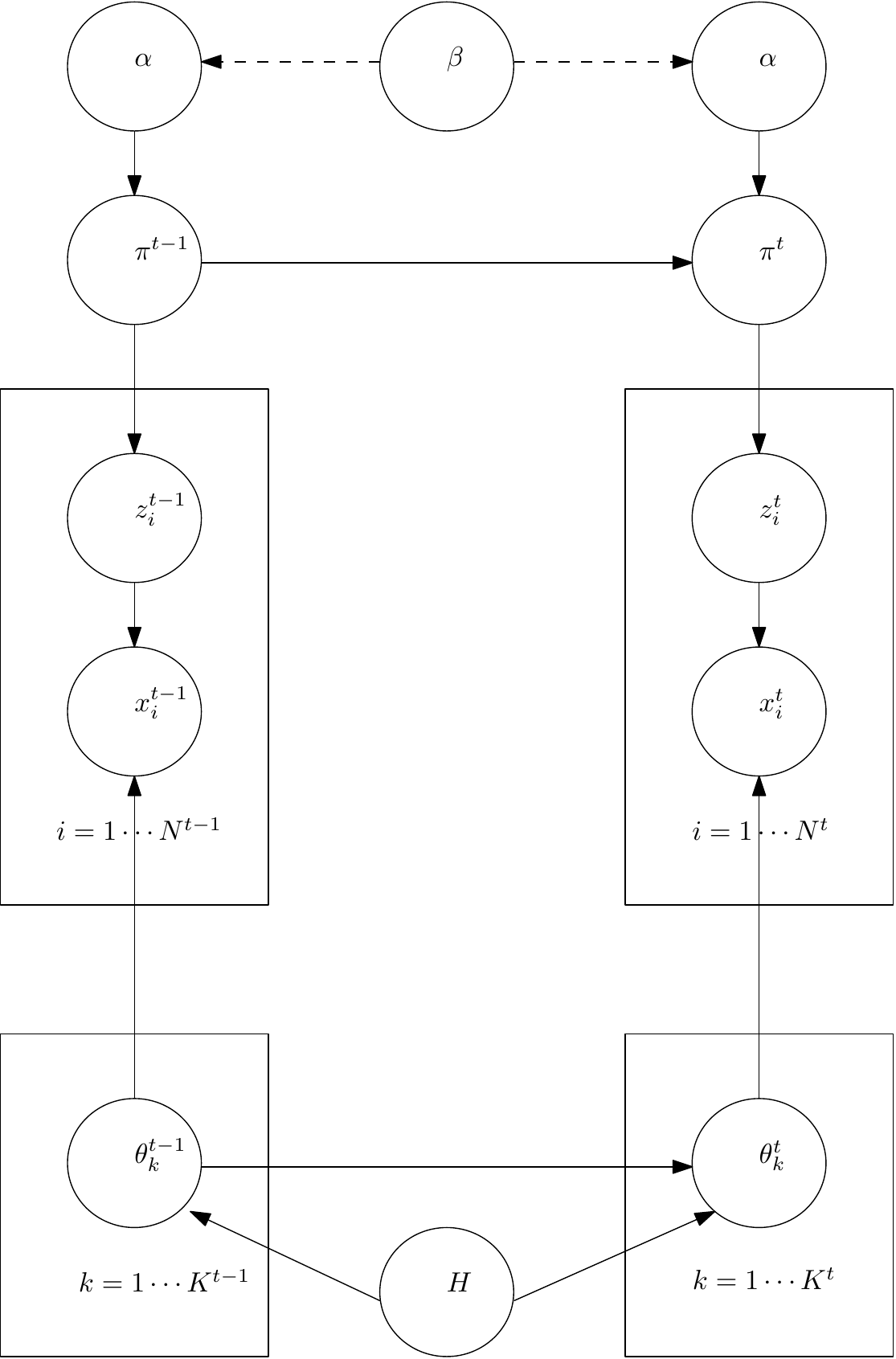}}
  \caption{ Model Evolution. Here, the dashed line represents the deterministic relation between $\beta$ and $\alpha$, where $\alpha = e^{-\beta}$. }
  \label{Fig:DPMM}
\end{figure}

Here, $x_i (i = 1 \cdots N)$ corresponds to the data and the $z_i(i = 1 \cdots N)$ corresponds to the latent variable, representing  cluster labels, taking one of the values from $k = 1 \cdots K$, where $N$ is the number of data points and $K$ is the number of clusters. $\pi$ is a vector of length $K$. $\pi_k$ represents the mixing proportion of data among the clusters, or the probability of $z_i$ taking the value $k$. $\theta_k$ is the parameter of the cluster $k$ and $F(\theta_{z_i})$ denotes the distribution defined by $\theta_k$. First, we pick $z_i$ from a Discrete distribution given in~(\ref{equation:DPM1}) and then generate data from a distribution parameterized by $\theta_{z_i}$ as given in~(\ref{equation:DPM2}), where the parameter $\pi$ is derived from a Dirichlet distribution as given in~(\ref{equation:DPM3}) and $\theta_k$ is derived from distribution $H$ of priors as represented in~(\ref{equation:DPM4}). The model is graphically~\cite{GraphicModel} presented in Fig.\ref{Fig:DPMM} (a).


Existing work~\cite{30,24,Neiswanger,Kuettel,IncrementalDPMM} do not emphasize much on how to come up with a suitable value of the concentration parameter for a given application. This paper derives a relationship between concentration parameter and distance function. Moreover, the method uses the distance information for associating the pixels to the same object on successive frames, thus addressing the temporal association of pixels to a cluster. This way, we are able to address both spatial and temporal dependencies of pixels belonging to the same object and thus makes it an the ideal choice for clustering pixels for segmentation and tracking applications. 

Our experiments reveal that, a single iteration of Gibbs sampling~\cite{25} can be sufficient to associate majority of the pixels that ensemble an object-in-motion to a single cluster. The cluster association can be maintained as long as the objects remain in motion. This concept can be extended for feature based clustering or segmentation, as distance between the pixels having similar characteristics is expected to be small. Thus, they can be grouped into a single cluster. We refer to this as Temporal Unknown Incremental Clustering (TUIC) model and the framework can be used to built tracking and surveillance applications. 

\subsection{Research Contributions}
This paper presents an incremental and hierarchical way of associating pixels to objects. Since the label of an object is maintained throughout its lifetime within the scene, this can further be extended hierarchically to derive most frequently used paths of the moving objects. Thus, we develop a framework that can be used during online detection of abnormal activities in surveillance videos. The main contributions are summarized as follows:
\begin{itemize}
\item A distance-based method for associating pixels to a cluster considering spatial as well as temporal properties of the moving objects.
\item We propose a method for deriving the concentration parameter ($\alpha$) in a given context or application.
\item Critical analysis of $\alpha$ during spatio-temporal segmentation of objects in various contexts, e.g. moving car surveillance, human motion analysis, outdoor surveillance, etc.
\item We propose a Temporal Unknown Incremental Clustering (TUIC) model for deriving an incremental learning framework that can be used for real-time activity detection.
\end{itemize}

The rest of the paper is organized as follows. In Section~\ref{Methodology}, proposed methodology is discussed. Section~\ref{sec:Results} presents experimental setup, dataset, parameters, analysis of results, and a few limitations of the proposed method. Section~\ref{sec:conclusion} concludes our work with discussion on a few future direction of the present work. 
\section{Proposed Methodology}
\label{Methodology}
\subsection{Background}
\label{sec:Background}
First, we discuss the terminologies used in the paper. Observation and data are used interchangeably. They represent words in text corpora or pixels in video frames. Similarly, we refer cluster or topic to represent distribution of data. A model is a representation of a real-world phenomenon. Model can be parametric or non-parametric. A parametric model is a family of distributions that can be described using finite set of parameters. Parametric model has a fixed number of parameters, while the number of parameters grow with the increase of training data for non-parametric model. A mixture model is a probabilistic model for representing the presence of sub-populations within an overall population. Mixture models can be finite or infinite. A finite mixture model is a probabilistic model representing a distribution of data from finite number ($K$) of sub-populations represented using a finite set of probability distributions. As $K \rightarrow \infty$, we get infinite mixture models. The Dirichlet distribution is the generalization of Beta distribution for multiple outcomes. A Dirichlet Process (DP)~\cite{ferguson1973} is a distribution over probability distributions and is used in Bayesian  non-parametric models, particularly in infinite mixture models known as Dirichlet Process Mixture Models (DPMM). Latent variables are variables that are not directly observed, rather inferred  from other observed variables. We use Graphical Model~\cite{GraphicModel} for representing mixture models. A graphical model is a probabilistic model for which a graph expresses the conditional dependence structure between the random variables. Random variables are represented by circles. Boxes are plates representing replicates. A graphical model represents the generative model of the data. 

 \subsection{Proposed Object Model}
\label{sec:Problem_formulation}
 In this paper, we use observation or data to represent pixel belonging to an object(s)-in-motion. Topic or cluster is used to represent the objects. Temporal segment or trajectories represent the tracks of the object(s)-in-motion. Any video frame can be modeled as a distribution of pixels belonging to objects and background.

\begin{figure}[!tbp]
  \centering
      \includegraphics[scale=0.48]{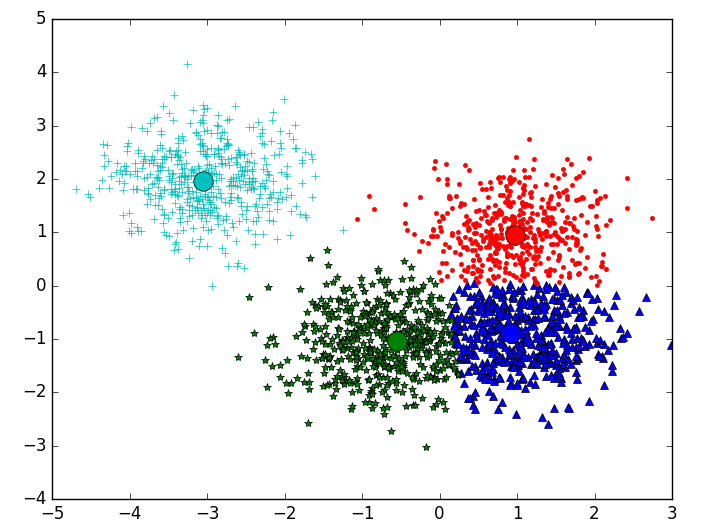}
    \caption{A Gaussian mixture representing four components. The centers are marked with circle.}
  \label{Fig:MOG}  
\end{figure}

 We have the following hypothesis to apply our proposed model for vehicular traffic analysis: 
 \begin{itemize}
\item[(i)] Since vehicles are rigid objects, all pixels belonging to a vehicle go through similar motion.
\item[(ii)] Size of the vehicular object in the image frames do not vary significantly within a short duration (between consecutive frames).
\item[(iii)] We assume that, the videos are captured from top view (near top view) using a static camera. Under normal circumstances, the vehicular objects present in an image frame are expected to be  spatially separated, i.e. they do not overlap.  
\end{itemize}
Here, we illustrate the rationale for building the model from a different perspective. Unlike the mixture of Gaussian as presented in Fig.~\ref{Fig:MOG}, the vehicles are rigid body objects.  The above assumptions make the problem simple as it indicates that the observations strictly belong to one topic. In addition to that, the pixels in the rigid body go through similar motion, i.e. they have similar magnitude and direction.

Let $x_i$ be the random variable representing the $i^{th}$ observation, where $i = 1 \cdots N$. $z_i$ is a discrete latent variable representing the cluster label. It can take values from ${1 \cdots K}$. Therefore, there are $N$ observations in the frame corresponding to $K$ clusters that represent the object(s)-in-motion. We want all the observations belonging to an object(s)-in-motion to be labeled correctly. Our goal is to find $z_i$ for all pixels.  $z_i$ corresponds to a discrete distribution and each $k \in \{1 \cdots K\}$ has a set of observations associated with it. The $k^{th}$ cluster has a proportion of observations. It is represented by $\pi_k$ that satisfies $\Sigma_{k=1}^{K} \pi_k = 1$. It can be observed that, $x_i$ associated with a particular cluster having unknown distribution parameterized by $\theta_k = \{\mu_k, \Sigma_k\}$, where $\mu_k$ and $\Sigma_k$ represent mean and covariance of the distribution. Let an observation be represented by \{$x$, $y$, $\Delta x$, $\Delta y$\}, where ($x$,$y$) represents the coordinate of the pixel in motion. $\Delta x$ and $\Delta y$ represent $x$ and $y$ components of the motion vector of the pixel. We build the inference scheme from the Bayesian representation of posterior as given in (\ref{equation:5}). 

\begin{equation}
\mbox{posterior} \propto \mbox{likelihood} \times \mbox{prior}
\label{equation:5}
\end{equation}

We can rewrite it in the current context using~(\ref{equation:6}), where $\mbox{likelihood}_{k_i}$ denotes the likelihood of $x_i$ with cluster label $k$ and $_k$ represents the prior probability of cluster $k$. We know that, prior$_k = \pi_k$. 

\begin{equation}
p(z_i = k | z_{-i},x_{-i},\theta_k) \propto  \mbox{likelihood}_{k_i} \times \mbox{prior}_k 
\label{equation:6}
\end{equation}

The above equation cannot be used to build the inference process as every parameter except $x_i$, is unknown. However, it gives a clue to derive the likelihood and prior required in the inference process. We build the clusters incrementally by considering observations one at a time since no information is available at the beginning. If we take the first observation, it forms a new cluster. $z_{-i}$ denotes the set of cluster assignment done so far, excluding that for the $i^{th}$ observation. Now, in subsequent observations, for example $x_i$, it is assumed that all observations sampled so far are assigned a cluster label. Using this information, the cluster label ($z_i$) for $x_i$ is found. $K$ denotes the number of clusters formed so far. Thus, initially $K = 0$ and as the sampling process progresses, it produces the actual number of clusters.  $\theta_{k_{-i}}$ denotes the parameter of cluster $k$ excluding the $i^{th}$ observation. This can be calculated since $x_{-i}$ and $z_{-i}$ are known.  Similarly, $n_{k_{-i}}$ denotes the number of observations present in $k^{th}$ cluster excluding the $i^{th}$ observation. The above equation is split to find the probability of $x_i$ to find new cluster label or an existing cluster label as given in~(\ref{equation:7}) and~(\ref{equation:8}), where $\pi_{k_{-i}}$ and $\pi_{{new}_{-i}}$ represent the prior probabilities of existing cluster $k$ and the $new$ cluster respectively.

\begin{equation}
p(z_i = new | z_{-i}, \theta_{k_{-i}}, \alpha) \propto \mbox{likelihood}_{k_i} \times \pi_{k_{-i}}
\label{equation:7}
\end{equation} 
\begin{equation}
p(z_i = k | z_{-i}, \theta_{k_{-i}}, \alpha) \propto \mbox{likelihood}_{new_i} \times \pi_{{new}_{-i}}
\label{equation:8}
\end{equation}  

Since the prior satisfies the property $\Sigma (\pi_{k_{-i}}) + \pi_{{new}_{-i}} = 1 $, prior for new cluster can be written as $\pi_{{new}_{-i}} = \frac{\alpha}{n_{-i} + \alpha}$ and for the $k^{th}$ cluster as $\pi_{k_i} = \frac{n_{k_{-i}}}{n_{-i} + \alpha}$, where $\alpha$ is the concentration parameter and $n_{-i} = \Sigma n_{k_{-i}}$. Here, $\alpha$ decides the probability of an observation forming a new cluster. $n_{-i}$ denotes the number of observations handled so far excluding $i^{th}$ observation. 
 Therefore, the problem is reduced to finding the likelihood function. We have described earlier in Fig.~\ref{Fig:OpticalFlow}(d), probability of pixel $p_1$  belongs to car \#1 is higher than that of pixel $p_2$. Similarly for pixel $p_2$, the probability of it belongs to car \#2 is higher than that of car \#1. It gives a visual clue about the property of the likelihood function, i.e. the probability is inversely proportional to the distance ($x$) of the pixel from the center of the object. Moreover, the probability is close to $0$ beyond the periphery of the object. An exponential decay function of the form $e^{-f(x)}$  satisfies the above property. The function also satisfies the condition $f(x) = 0$ for $x_i$ to form a new cluster as distance to itself is 0. Thus, likelihood function of a new cluster is $1$. By taking $new = K+1$, we rewrite~(\ref{equation:7}) and~(\ref{equation:8})  by~(\ref{equation:9}) and~(\ref{equation:10}).

\begin{equation}
p(z_i = K+1 | z_{-i}, \theta_{k_{-i}}, \alpha) \propto \frac{\alpha}{n_{-i} + \alpha}
\label{equation:9}
\end{equation} 
\begin{equation}
p(z_i = k | z_{-i}, \theta_{k_{-i}}, \alpha) \propto \text{ }e^{-f(x)} \times \frac{n_{k_{-i}}}{n_{-i} + \alpha}
\label{equation:10}
\end{equation}  

We can further simplify the equation as the denominator of prior is constant at any sampling point. $\alpha$ = $e^{-\beta}$ is assumed to represent the inference equations in similar form, i.e. in terms of number of observations and exponential decay function. The proportionality symbol is removed by introducing a normalization constant $b$. Now, a generalized formula is given in~(\ref{equation:11}). This equation is the key to find the value of the concentration parameter as $\beta$ can be expressed in terms of distance from the center to a point at periphery of the object. The method is discussed in Section~\ref{sec:Results}. 

\begin{align} \label{equation:11}
p(z_i = k | z_{-i},x_{-i},\theta_{k_{-i}},\beta)
&= \begin{cases}
b \times e^{-\beta},        &\text{if $k = K + 1$;}\\
b \times e^{-f(x)} \times n_{k_{-i}},  &\text{otherwise.}
\end{cases}
\end{align} 

\begin{figure}[!tbp]
  \centering
       \includegraphics[scale=0.48]{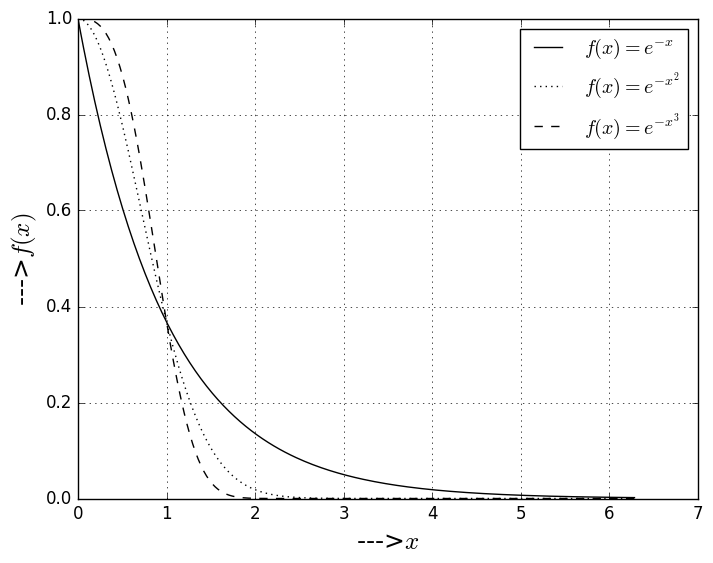}
    \caption{Examples of a few exponential decay functions.}
  \label{Fig:EDecay}  
\end{figure}

$f(x)$ can be assumed to be in the form of $ax^m$, where $a$ is a constant and $m$ is the exponent. A few decay functions with $a = 1$ are shown in Fig.~\ref{Fig:EDecay} for 
$m = \{1, 2, 3\}$ assuming Euclidean distance. Now, we can rewrite the above equation using~(\ref{equation:12}).

\begin{align} \label{equation:12}
p(z_i = k | z_{-i},x_{-i},\theta_{k_{-i}},\beta)
&= \begin{cases}
b \times e^{-\beta},        &\text{if $k = K + 1$;}\\
b \times e^{-ED} \times n_{k_{-i}},  &\text{otherwise.}
\end{cases}
\end{align} 

 It can be observed that, when square of the Euclidean distance is used to compute $f(x)$ in case of Multivariate Gaussian, it becomes a special case of the distance function $f(x)$.  $(x_i - \mu)^T \Sigma^{-1} (x_i - \mu)$ is a special case of squared Euclidean distance for Multivariate Gaussian, where $\Sigma$ is the covariance matrix for the cluster and $\mu$ the mean of the cluster. Hence the relation given in~(\ref{equation:12}) can be written as ~(\ref{equation:13}) and~(\ref{equation:14}), where $\mu_{-i}$ is the mean of the distribution.   
     

\begin{equation}
p(z_i = K + 1 | z_{-i}, \theta_{k_{-i}}, \beta) = b \times e^{\beta}
\label{equation:13}
\end{equation} 
\begin{equation}
p(z_i = k | z_{-i}, \theta_{k_{-i}}, \beta) = b \times e^{-\sqrt{((x_i - \mu_{-i})^T (x_i - \mu_{-i}))}} \times {n_{k_{-i}}}
\label{equation:14}
\end{equation} 

The formulas given in~(\ref{equation:13}) and~(\ref{equation:14}) represent the inference equations of our object model. The object Model can be represented as shown in Fig.~\ref{Fig:DPMM} (b). This forms the basis of our proposed Temporal Unknown Incremental Clustering (TUIC) model described next.
 
\subsection{Temporal Unknown Incremental Clustering (TUIC) Model}
\label{sec:Problem_formulation_2}
We extend the assumptions about the motion of objects across video frames. We build our model based on the following additional assumptions:
\begin{itemize}
\item[(i)]The objects do not move substantially between successive frames, hence there will be overlap between pixels belonging to object(s)-in-motion between successive frames.
\item[(ii)]The object motion features do not change significantly between $t-1$ and $t$, where $t$ represent the time stamp of the frame, i.e. state information does not change significantly between frames.
\end{itemize}     

If $i^{th}$ pixel belongs to an object in both $(t-1)^{th}$ and ${t}^{th}$ frames, the probability of an observation $x_i^{t}$ belongs to a cluster $z_i^{t-1}$ is  expected to be higher than it belongs to other clusters. This implies, cluster parameters are approximately equal between successive frames, i.e. $\theta_k^{t}  \approx \theta_k^{t-1} $. However, they may not be exactly same. If Gibbs sampling is performed using  $\theta_k^{t-1}$ as a prior for the ${t}^{th}$ frame, not only the convergence becomes faster, but also the cluster labels are maintained between consecutive frames. The inference can be done as per~(\ref{equation:17}) and~(\ref{equation:18}) with exactly one iteration of the Gibbs sampling. The rationale behind using only one iteration per frame is that, even if all the observations do not get clustered correctly in the current frame, they are essentially done in the subsequent frames since the features do not change significantly between consecutive frames. Here, $z_{-i}^*$ is different from the $z_{-i}$ discussed earlier. It represents the set of all cluster assignments except for $x_i^{t}$ such that it includes only the latest elements between $z_i^{t-1}$ and $z_i^{t}$ for any $i$. $\theta_{k_{-i}}^*$ is the parameter representing the distribution corresponding to cluster $k$ in time-stamp $t$ from the set of observations corresponding to $z_{-i}^*$, where $\mu_{-i}^*$ is the mean of $\theta_{k_{-i}}^*$ distribution. $n_{k_{-i}}^*$ denotes the number of observations in $\theta_{k_{-i}}^*$ and $b$ is the normalization constant. Our proposed model is represented in Fig.~\ref{Fig:DPMM} (c) as a generative model and can be represented using ~(\ref{equation:ITCM1}-\ref{equation:ITCM4}). 
   
\begin{equation}
p(z_i^{t} = K+1 | z_{-i}^*, \theta_{k_{-i}}^*, \beta) = b \times e^{-\beta}
\label{equation:17}
\end{equation} 
\begin{equation}
p(z_i^{t} = k | z_{-i}^*, \theta_{k_{-i}}^*, \beta) = b \times n_{k_{-i}}^* \times e^{\sqrt{(x_i^{t}  - \mu_{-i}^* )^T (x_i^{t} - \mu_{-i}^*)}}
\label{equation:18}
\end{equation} 
\begin{equation}
z_i^t|\pi^t  \sim  \mbox{Discrete}(\pi^t)
\label{equation:ITCM1}
\end{equation} 
\begin{equation}
x_i^t|z_i, \theta_{z_i^t}  \sim  F(\theta_{z_i^t})
\label{equation:ITCM2}
\end{equation} 
\begin{equation}
\pi^t = (\pi_1^t, \cdots ,\pi_K^t)|e^\beta,\pi^{t-1}  \sim  \mbox{Dirichlet}(e^\beta / K^t, \cdots, e^\beta / K^t)
\label{equation:ITCM3}
\end{equation} 
\begin{equation}
\theta_k^t|H,\theta_k^{t-1}  \sim  H
\label{equation:ITCM4}
\end{equation}

Here, $x_i^t (i = 1 \cdots N)$ corresponds to the data at time $t$ and $z_i^t(i = 1 \cdots N)$ corresponds to the latent variable representing  cluster labels, taking one of the values from $k = 1 \cdots K^t$. $N$ is the number of data points and $K^t$ is the number of clusters. $\pi^t$ is a vector of length $K^t$. $\pi_k^t$ represents the mixing proportion of data among clusters. $\theta_k^t$ is the parameter of the cluster $k$ and $F(\theta_{z_i^t})$ denotes the distribution defined by $\theta_k^t$. First, we pick $z_i^t$ from a Discrete distribution given in~(\ref{equation:ITCM1}). The data is then generated from a distribution parameterized by $\theta_{z_i}^t$ as given in~(\ref{equation:ITCM2}), where the parameter $\pi^t$ is derived from a Dirichlet distribution as given in (\ref{equation:ITCM3}). $\theta_k^t$ is derived from another distribution $H$ of prior as represented in~(\ref{equation:ITCM4}). It may be observed that, the model is different from the original DPMM shown in Fig.~\ref{Fig:DPMM} (a). In the original model, there is a conditional dependence between $\theta_k^t$ and $\theta_k^{t-1}$, or $\pi^t$ and $\pi^{t-1}$.

Inference method for cluster assignment uses Gibbs sampling~\cite{25}. The process is described in Algorithm~\ref{algorithm:1}. Firstly, optical flow ~\cite{Farnebäck2003} is extracted. A data point $x_{i}$ is denoted by the 4-tuple ($x$, $y$, $\Delta x$, $\Delta y$), where ($x$, $y$) represent the coordinates of the pixel and $\Delta x$ and $\Delta y$ represent x and y components of optical flow vector of the pixel. A threshold has been applied on the magnitude of the optical flow to remove the pixels not having any optical flow. This has been done purposefully to categorize pixels belonging to the background to a single cluster. However, existing background detection methods can be used to push pixels which are irrelevant for clustering. 

\algnewcommand\algorithmicswitch{\textbf{switch}}
\algnewcommand\algorithmiccase{\textbf{case}}
\algnewcommand\algorithmicassert{\texttt{assert}}
\algnewcommand\Assert[1]{\State \algorithmicassert(#1)}%
\algdef{SE}[SWITCH]{Switch}{EndSwitch}[1]{\algorithmicswitch\ #1\ \algorithmicdo}{\algorithmicend\ \algorithmicswitch}%
\algdef{SE}[CASE]{Case}{EndCase}[1]{\algorithmiccase\ #1}{\algorithmicend\ \algorithmiccase}%
\algtext*{EndSwitch}%
\algtext*{EndCase}%

\begin{algorithm}[!h]
\caption{Temporal Unknown Incremental Clustering (TUIC)}
\textbf{Input:} Input video, $\beta$ \\
\textbf{Output:} Labelled video
\begin{algorithmic}[1]
\State Initialize a background cluster $c_0 = (0,0,0,0,0,0)$, where $c_k$ is a random variable representing 6-tuple $(\mu_x, \mu_y, \mu_{\Delta x}, \mu_{\Delta y}, z, t_{dur})$ corresponding to a cluster. $\mu$ represents the mean value of respective parameters and $z$ takes $k = \{1 \cdots K$\}. $t_{dur}$ represents the time duration of the cluster label $k$;
\State Initialize $o_i = (x,y,0,0,0)\; \forall\: i={1 \cdots N}$, where $o_i$ represents a 2-tuple ($x_i$, $z_i$) containing pixel and label information. Add $o_i$ to $c_0$  $\forall i$;
\State $K = 0$;
\State Get the next frame; 
\For {frame(!Empty)}
\State Get optical flow and fill $x_i \in o_i \; \forall i$;
\For{each $i$ }
\State Remove $o_i$ from $c_k$; 
\State Find $z = z_i^{t}$ corresponding to MAX [$p(z_i^{t} = K+1 | z_{-i}^*, \theta_{k_{-i}}^*, \beta)$, $p(z_i^{t} = k | z_{-i}^*, \theta_{k_{-i}}^*, \beta)$] as given in~(\ref{equation:17}) and~(\ref{equation:18}), respectively;
  \Switch{($z$)}
    \Case{$K + 1$:}
      \State $K = K + 1$;
      \State Create a new cluster $c_K$;
      \State Set $z_i = K$ in $o_i$;
      \State Update cluster parameters ($\mu_K$, $\Sigma_K$);      
    \EndCase
    \Case{$k$:}
      \State Set $z_i = k$ in $o_i$;
      \State Update cluster parameters ($\mu_k$, $\Sigma_k$);
    \EndCase
  \EndSwitch
\EndFor
\State Display the frame with cluster labels;
\State Get the next frame;
\EndFor
\end{algorithmic}
\label{algorithm:1}
\end{algorithm}
  
\section{Results and Discussions}
\label{sec:Results}
In this section, we describe experiments conducted using the proposed TUIC model and results obtained using various public datasets. We also present comparative analysis with state-of-the-art techniques. 

\subsection{Experimental Setup and Datasets}
\label{sec:experimental_setup}
\begin{figure}[!tbp]
  \centering
      \includegraphics[width=0.7\columnwidth]{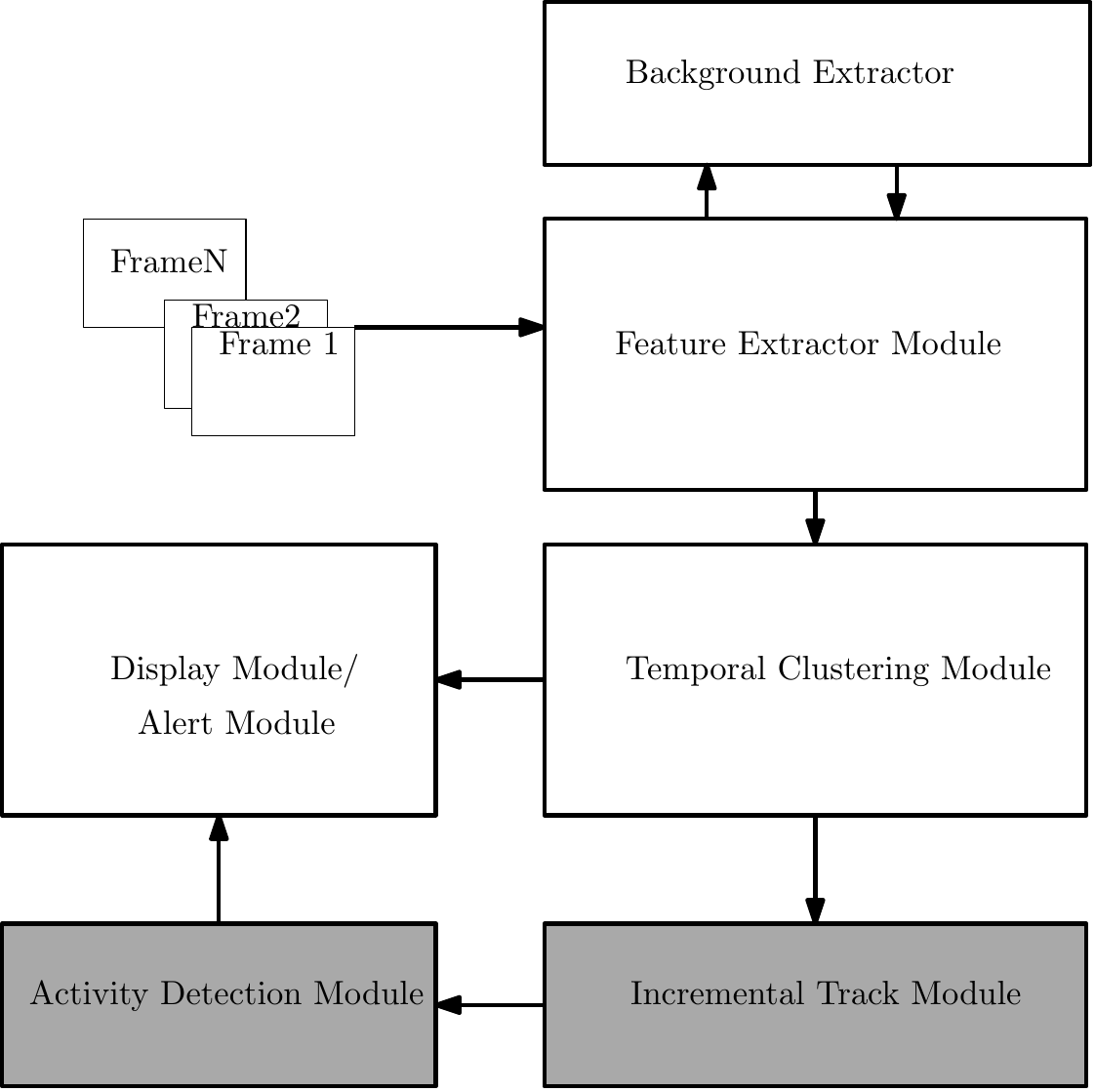}
    \caption{Proposed traffic analysis framework.}
  \label{Fig:Framework}  
\end{figure}

We have used OpenCV to implement our proposed framework. Experiments have been conducted on three publicly available traffic datasets, namely VIRAT, MIT, and UCF as mentioned in~\cite{Dataset}. Fig.~\ref{Fig:Framework} depicts the complete framework of our implementation. Optical flow is calculated from successive video frames using Farne-Back method~\cite{Farnebäck2003}. A background separation module has been used to extract the motion pixels or foreground. We quantize the motion at each foreground pixel in eight directions. Background pixels are static with no motion in any direction. Motion of the foreground pixels are then considered to construct the model. The blocks highlighted in grey are kept for future extensions required to build a complete traffic analysis framework. Incremental tracking module takes the trajectories or tracklets (${t}_k =<x_{start},y_{start}>,....,<x_{end},y_{end}>$) generated using the proposed TUIC model. They can also be used to find most frequently used segments of a road. Here, $<x_{start},y_{start}>$ and $<x_{end},y_{end}>$ represent start and end positions of the cluster. Activity detection module can be used for detecting interesting activities from the learned model. 

\subsection{Empirical Evaluation of $\beta$}
\label{sec:Parameter}
TUIC model has one parameter ($\beta$) that can influence the results of the inference process. \textbf{$\beta$} is referred to as the negative exponent of concentration parameter of the model and it decides the size of the object to be traced. In~(\ref{equation:11}), a basis for setting a suitable value of this parameter is presented. If an observation forms a new cluster, it has to be at a distance higher than that of the object periphery. This implies the relation given in~(\ref{eq:beta_init}) holds true. 
\begin{equation}
\label{eq:beta_init}
e^{-\beta} = e^{-f(x + \delta x)} \times n_{k_{-i}}
\end{equation}
Therefore, value of $\beta$ can be estimated using~(\ref{eq:beta_initial})
\begin{equation}
\label{eq:beta_initial}
\beta =  f(x + \delta x) - \ln(n_{k_{-i}}).
\end{equation}
Now, if we use maximum distance ($max$) to the periphery of the object and number of observations ($n_k$) in an object is known, (\ref{eq:beta_initial}) can be rewritten as~(\ref{equation:beta1})
\begin{equation}
\label{equation:beta1}
\beta =  f(max + \delta x) - \ln(n_k).
\end{equation} 

\begin{figure*}[h]
\centering
\subfigure[Euclidean Distance v/s time-stamp]{\includegraphics[scale=0.34]{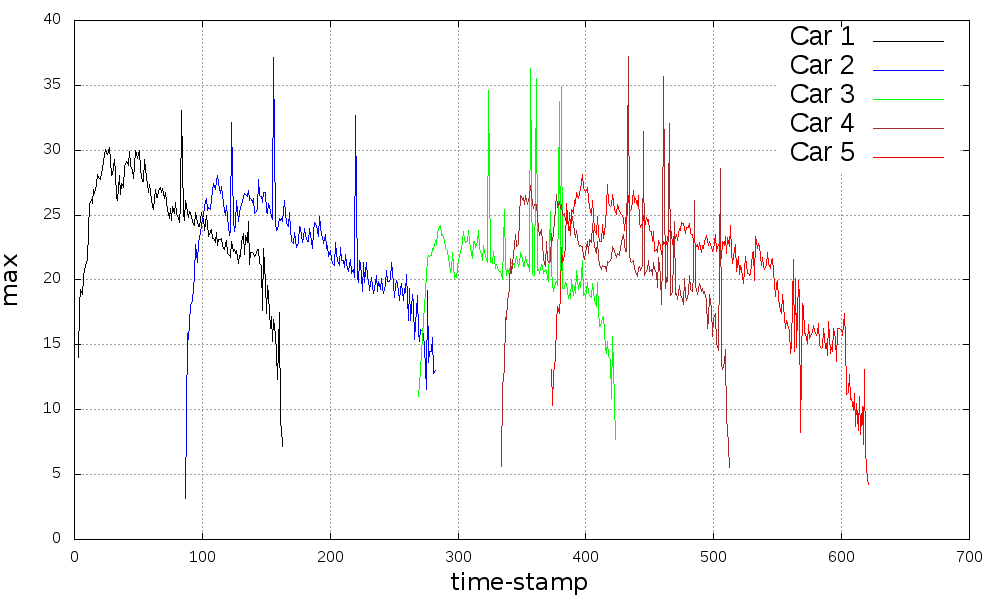}} 
  \subfigure[Number of pixels v/s time-stamp]{\includegraphics[scale=0.34]{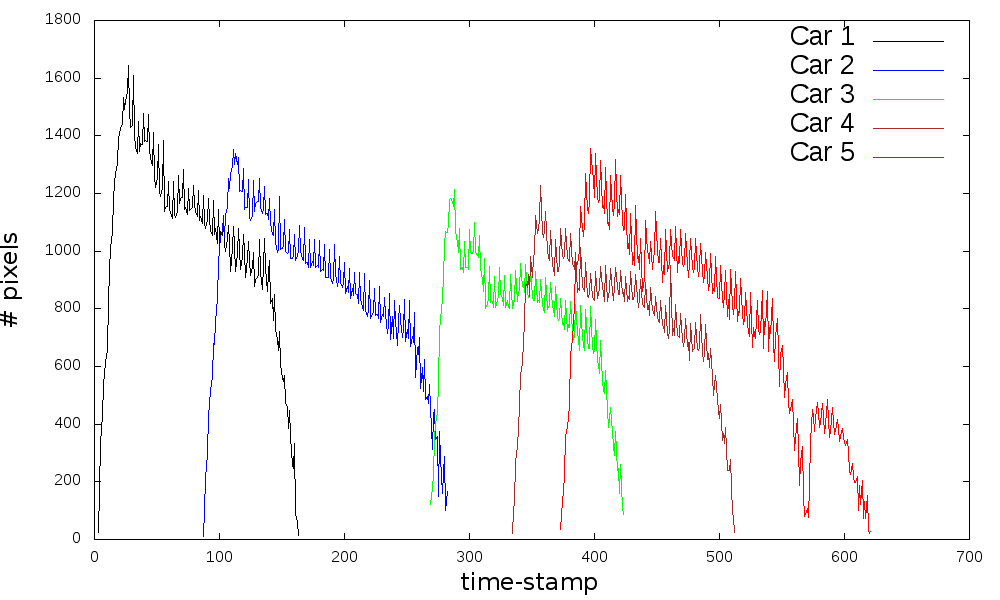}}   
  \caption{Critical analysis of the distance function to estimate $\beta$. It can be observed that the relation between the maximum distance to the periphery and the number of observations present in the clusters are linearly related. This justifies the model taking the distance as a measure for doing the clustering.}
  \label{Fig:CriticalAnalysis}
\end{figure*}

Initially, we set approximate values of $n_k$ and $\delta x$ by calculating the distance from the center of the object to its periphery and run the clustering algorithm on a video with single object. If more than one clusters are formed corresponding to the object, we increase $max$ or decrease $n_k$ to obtain single cluster corresponding to the object. Finally, we get the actual values of $n_k$ and $\delta x$. Values of $n_k$ and $\delta x$ can then be used for calculating $\beta$. Fig.~\ref{Fig:CriticalAnalysis} shows how distance varies over time. $\beta$ can be estimated once the values of $n_k$ and $\delta x$ are known. However, even if we fix $\beta$, object size may not be fixed temporally. This is because of perspective view as the surveillance cameras are often installed to capture long-range views of the scene, they may not capture the top view always.

Our experimental study also reveals that majority of the objects  follow similar trajectory or pattern, thus traces of $max$ over time remain similar as depicted in Figs.~\ref{Fig:CriticalAnalysis} (a) and (b). It may be noted that the distance gradually increases as the objects enter the scene and reach a peak value which is maintained (or decreases slowly) for some time. Then it decreases suddenly as the objects move out of the scene. However, if top view recording can be obtained, it is expected that $\beta$ will remain flat for longer duration. It can also be observed that number of pixels forming a cluster (object) varies similarly. Therefore, a correlation between $max$ and number of pixels can be established.  The spikes in the curves are due to noisy observations getting added to the cluster which can be removed using appropriate filtering. 

\subsection{Variations in Sampling Orders}
Since the observations may have spatial dependency, they cannot be interchanged as proposed in the original Dirichlet Process~\cite{Farnebäck2003}. Therefore, we have carried out a set of experiments with different sampling orders as listed hereafter to understand the effect of sample ordering. 
\begin{itemize}
\item Linear Sampling: Pixels are sampled columnwise starting from the first to the last column. We then go rowwise.
\item Random Sampling: Pixels are sampled in a random order.
\item Spiral Sampling: Pixels are sampled in a spiral manner.
\end{itemize}
Our experiments reveal that despite variations in sampling ordering, they work fairly well to produce good trajectories with a suitable $\beta$. All sampling orders produce split clusters at some point of time due to the noises. In all sampling methods, there are issues while the objects leave the scene and they are in close proximity. Some cluster may get merged while approaching the boundary. However the cluster labels are maintained correctly till the objects exit the scene. 


\subsection{Selection of the Distance Function}
\begin{figure*}[h]
  \centering
  \subfigure[Frame~\#164]{\includegraphics[scale=0.38]{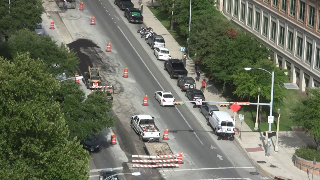}}
    \subfigure[likelihood = $e^{-x}$]{\includegraphics[scale=0.38]{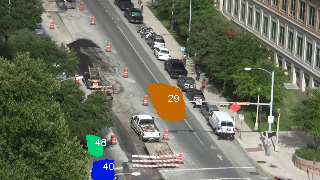}}  
    \subfigure[likelihood = $e^{-x^2}$]{\includegraphics[scale=0.38]{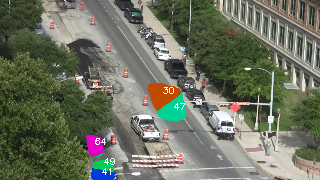}}  
    \subfigure[likelihood = $e^{-x^3}$]{\includegraphics[scale=0.38]{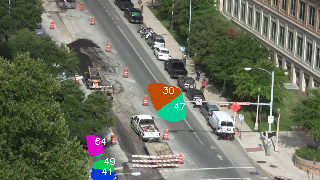}}   
\subfigure[Value of maximum distance ($\max$) for object~\#2]{\includegraphics[scale=0.34]{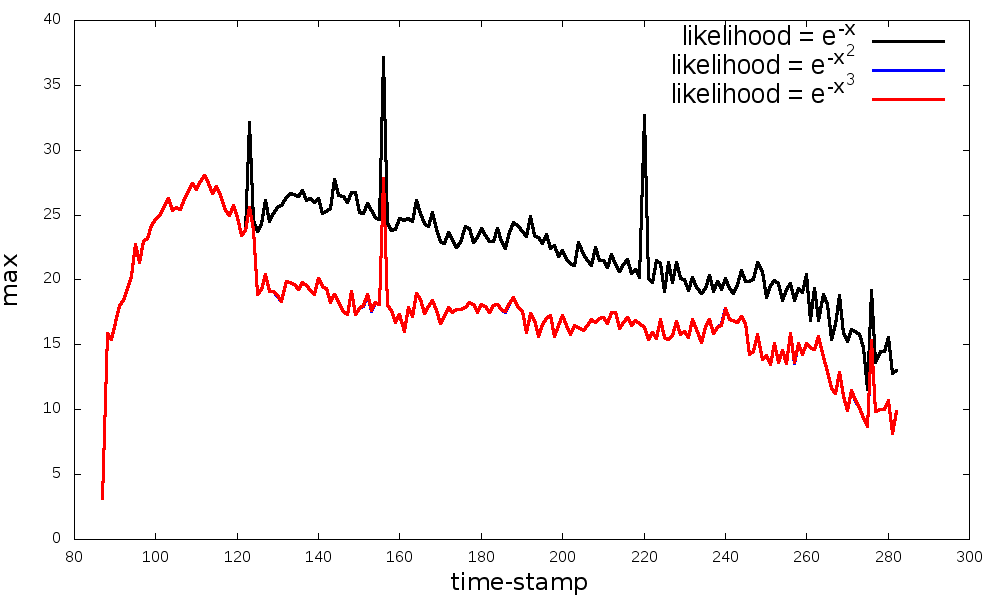}}  
\subfigure[Number of pixels for object~\#2]{\includegraphics[scale=0.34]{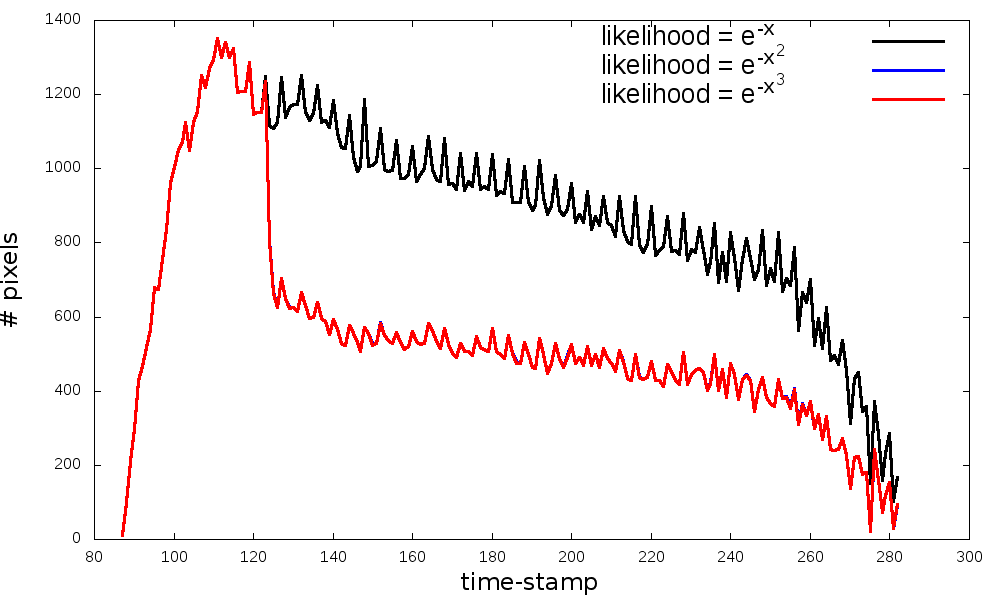}} 

  \caption{Clustering using different distance functions. $e^{-x}$ produces stable clustering for a fixed $\beta$ value as compared to other likelihood functions. Even though the objects are split into more than one clusters due to noises. A careful observation reveals that the likelihood functions $e^{-x^2}$ and $e^{-x^3}$ produces similar curves.}
  \label{Fig:DISTANCE_FUNC}
\end{figure*}

In Fig.~\ref{Fig:DISTANCE_FUNC}, we compare clustering results using three different distance functions, e.g. $f(x) = x, f(x) = x^2, f(x) = x^3$. The results reveal that the performance can vary. However, $f(x) = e^{-x^2}$ or$ f(x) = e^{-x^3}$ look quite similar as depicted Fig.~\ref{Fig:EDecay}. A smaller value of $\delta x$ makes sure that the relation between $max$ and $n_k$ remains linear. Therefore, we have used $f(x) = e^{-x}$ in our experiments.  According to our observation, the best results are obtained using $e^{-max}$ with $\delta x = 1$. 

\subsection{Noise Removal}
\begin{figure*}[h]
  \centering
  \subfigure[Original Noisy Clusters]{\includegraphics[scale=0.38]{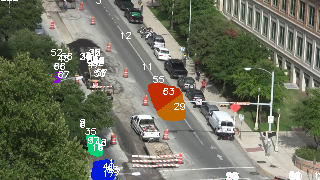}}
  \subfigure[Frame Skipping]{\includegraphics[scale=0.38]{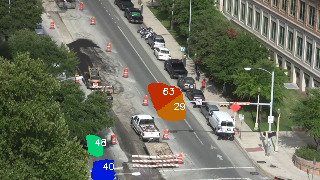}}  
  \subfigure[Moving Average Filtering]{\includegraphics[scale=0.38]{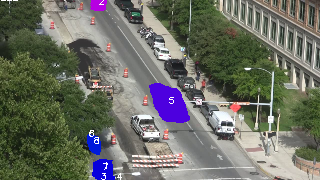}}
  \subfigure[Moving Average Filtering with Frame Skipping]{\includegraphics[scale=0.38]{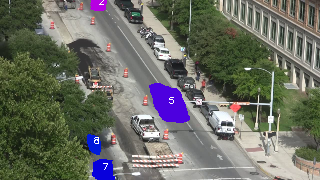}}  
  \caption{Effect of noises in clustering and removal of noises using frame skipping and moving average filter.}
  \label{Fig:Noise}
\end{figure*}

\begin{figure*}[h]
  \centering
  {\includegraphics[scale=0.5]{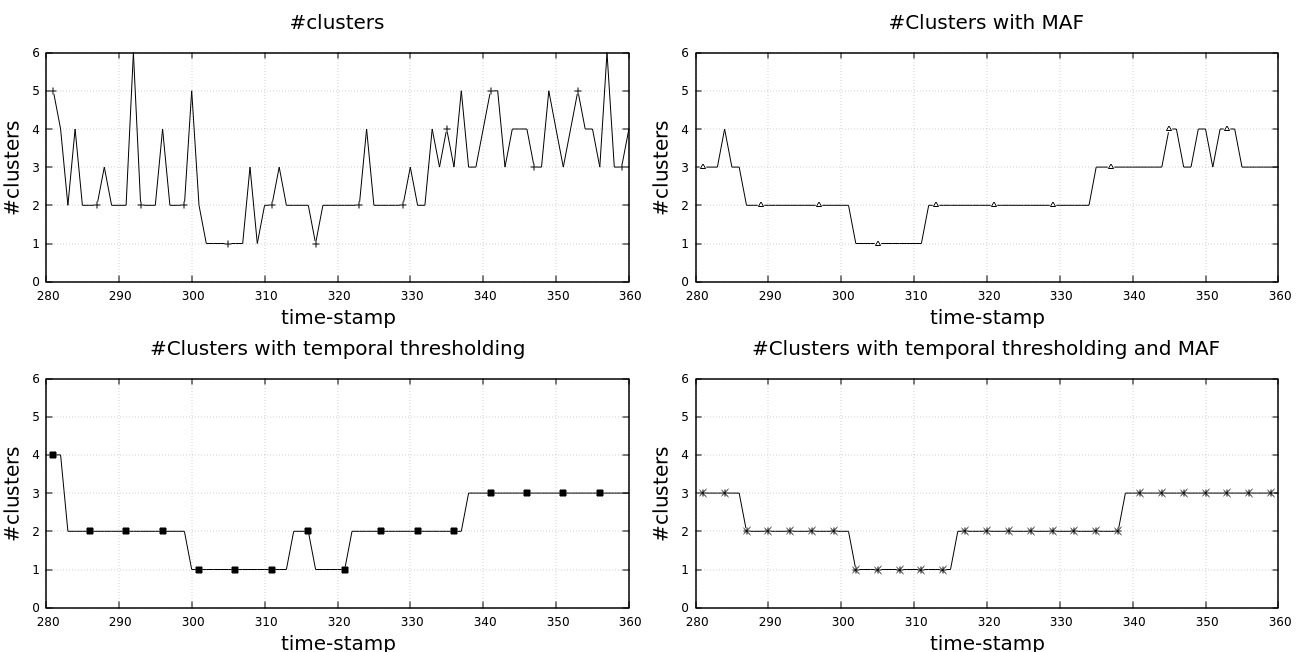}}
  \caption{Effect of noise removal on number of valid clusters.}
  \label{Fig:NoisePlot}
\end{figure*}

\begin{figure}[h]
  \centering
  
{
  \includegraphics[scale=0.35]{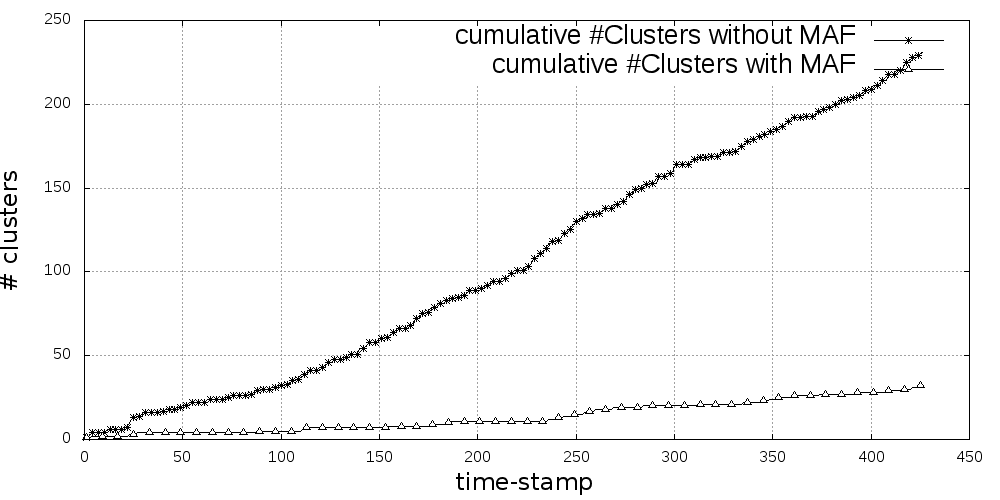}}    
  \caption{The effect of MAF on the number of clusters can be seen from the cumulative plot of the \#clusters over time.}
  \label{Fig:CumulativeNoisePlot}
\end{figure}

Optical flow corresponding to slowly moving objects may not be significant when the videos are processed at high frame rate. Also, we have observed that the proposed model can be used to track small objects such as humans by processing the videos at lower frame rate. However, results can be affected if the cluster representing an object does not remain live for at least three successive frames. Such clusters are referred to as noise. Clusters that are shown earlier have been selected since they lasted in more than three frames. Alternatively, a moving average filter (MAF) can be applied on the optical flow for smoother features between successive frames. Fig.~\ref{Fig:Noise} depicts the noise level and its impact on clustering. The role played by noise in clustering can be interpreted from Fig.~\ref{Fig:NoisePlot} and Fig.~\ref{Fig:CumulativeNoisePlot}. Since MAF with temporal thresholding gives best results, further experiments are conducted after noise removal using MAF.

\subsection{Experiments on Various Public Datasets}
\begin{figure*}[h]
  \centering
  \subfigure[Original frame]{\includegraphics[scale=0.3]{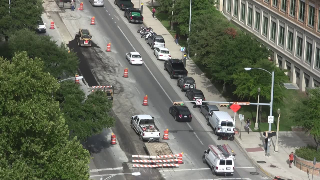}}  
  \subfigure[$\beta = 5$]{\includegraphics[scale=0.3]{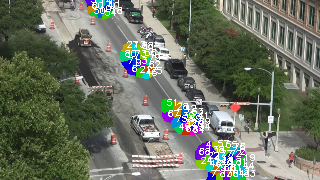}}
  \subfigure[$\beta = 15$]{\includegraphics[scale=0.3]{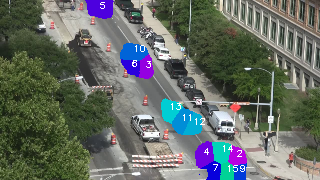}}
  \subfigure[$\beta = 35$]{\includegraphics[scale=0.3]{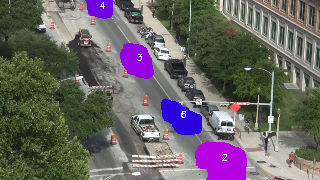}}
  \subfigure[$\beta = 155$]{\includegraphics[scale=0.3]{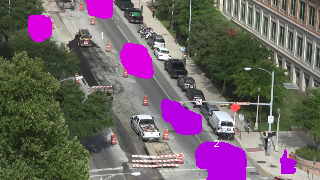}}  
  \caption{Impact of $\beta$ on the performance of the clustering. For smaller values of $\beta$, say 5 or 15, number of clusters per object have been found to be high. As $\beta$ increases, number of clusters per object reduces. Optimum clustering has been obtained at $\beta = 35$. More than one objects grouped into a single cluster when $\beta$ is higher.}
  \label{Fig:BETA}
\end{figure*}

We now present the experiment results obtained using various publicly available datasets. We have applied our proposed clustering 
in two different contexts, namely road traffic analysis and human motion analysis. In order to establish the relation of $\beta$ with the clustering process, a set of experiments have been carried out on VIRAT dataset videos and the results are presented in Fig.~\ref{Fig:BETA}. 
\begin{figure}[!tbp]
  \centering
\includegraphics[scale=0.6]{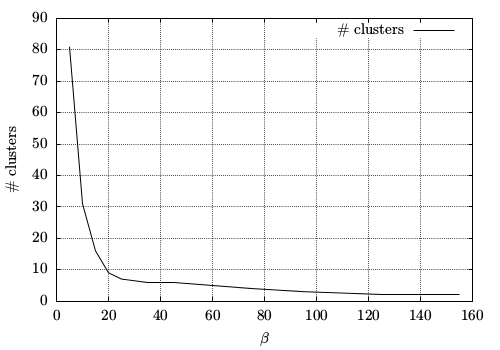}
    \caption{Variation in number of clusters with different $\beta$ for a particular frame.}
  \label{Fig:BetaVsClusters}  
\end{figure}

It has been observed that with a smaller value of $\beta$, more clusters are usually formed for any single moving object.  Fig.~\ref{Fig:BETA} depicts how clusters per object vary as $\beta$ varies. Our experiments reveal that more than one objects are merged into a single cluster when a larger value of $\beta$ is used. An example of this phenomenon is depicted in Fig.~\ref{Fig:BETA} (f). Graphical plot shown in Fig.~\ref{Fig:BetaVsClusters} depicts how the number of clusters vary within a frame when $\beta$ is varied.

\begin{figure*}[h]
  \centering
  \subfigure[Original Frame]{\includegraphics[scale=0.3]{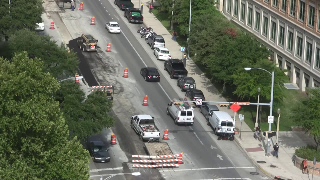}}  
  \subfigure[$\beta = 15$ Clusters]{\includegraphics[scale=0.3]{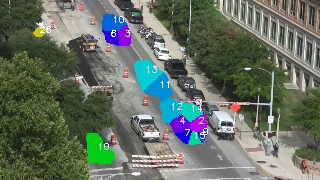}}
  \subfigure[$\beta = 15$ Trace]{\includegraphics[scale=0.3]{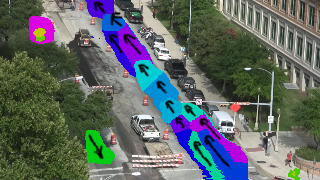}}
  \subfigure[$\beta = 35$ Clusters]{\includegraphics[scale=0.3]{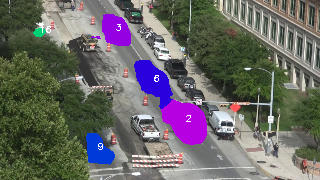}}
  \subfigure[$\beta = 35$ Trace]{\includegraphics[scale=0.3]{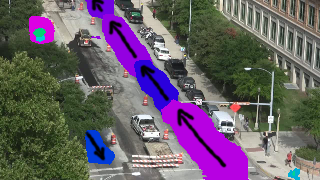}}  
  \caption{ The figures corresponds to frame \#45 of the video used for $\beta$ variation. In spite of $\beta$ betting varied the cluster labels are traced temporally as can be seen in (d) and (f).}
  \label{Fig:BETA_TRACE}
\end{figure*}

Another observation is, actual objects are smaller as compared to the clusters. This happens because the optical flow algorithm  estimates magnitude of the flow in neighborhood pixels. This implies, a better approximation of the object can be obtained with more accurate optical flow. We have maintained single cluster label throughout the life of the object as depicted in Fig.~\ref{Fig:BETA_TRACE}. If we choose a $\beta$ value corresponding to the smallest object, larger objects may be divided into more than one clusters. Therefore, post processing needs to be used to merge them. This way, our proposed model can be the basis for object motion analysis. For example, we can find whether the object-in-motion is a small vehicle/medium vehicle/large vehicle based on the number of clusters connected together.

\begin{figure}[h]
  \centering
  \subfigure[Frame marked with motion information]{\includegraphics[scale=0.20]{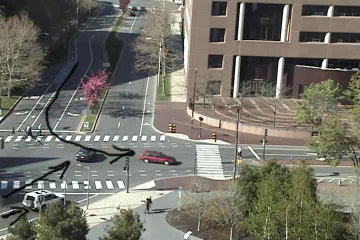}}
  \subfigure[Clusters formed on the frame]{\includegraphics[scale=0.20]{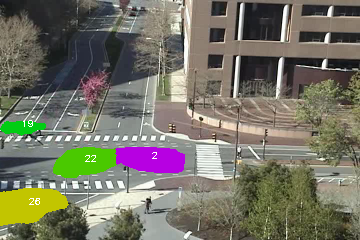}}
  \subfigure[Traces of the three objects]{\includegraphics[scale=0.20]{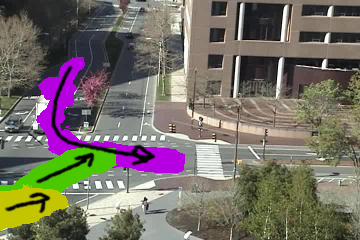}}
  \caption{Detection and tracking of three cars in MIT traffic dataset video.}
  \label{Fig:MIT_TRAFFIC}
\end{figure}

\begin{figure}[h]
  \centering
  \subfigure[Frame marked with motion path]{\includegraphics[scale=0.20]{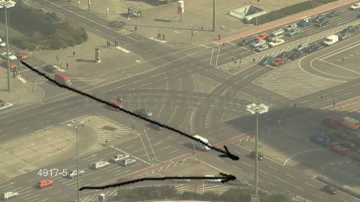}}
  \subfigure[Clusters formed on the frame]{\includegraphics[scale=0.20]{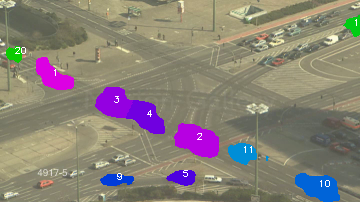}}
  \subfigure[Traces of the three objects]{\includegraphics[scale=0.20]{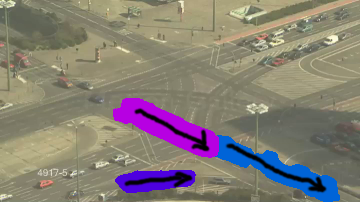}}
  \caption{Detection and tracking of moving objects in UCF traffic dataset video. The snapshot is on 50th frame where two signals started and corresponding traffic flows are marked.}
  \label{Fig:UCF_TRAFFIC}
\end{figure}

We have conducted tests on other public datasets, namely MIT, and UCF videos. Results reveal that the inference scheme is able to cluster the objects incrementally with good accuracy. Such results are presented in Fig.~\ref{Fig:MIT_TRAFFIC} and Fig.~\ref{Fig:UCF_TRAFFIC}. We have shown the traces of moving objects using different colors. Unlike VIRAT videos where the objects are fully tracked till the end as they do not occlude, some of the objects present in MIT and UCF videos are represented by more than one clusters. This happens due to a fixed $\beta$ value to facilitate clustering of medium-sized vehicles. Corresponding optical flow reveals that the vectors are overlapping. Moreover, there are vehicles of different dimensions. We can therefore use a smaller $\beta$ value to facilitate clustering of smaller vehicles. Post-processing has been used (connected components) to generate tracks. These accurate tracks can further be used for Motif analysis at traffic junctions without involving complex modeling as adopted in~\cite{2}. Our proposed clustering can be used to cluster trajectories for high-level information retrieval. The clusters with more number of tracks can form the frequently occurring trajectory patterns (motifs). 

\subsection{Experiments on Crowd Datasets}
\begin{figure}[h]
  \centering
  \subfigure[Frame marked with motion information]{\includegraphics[scale=0.22]{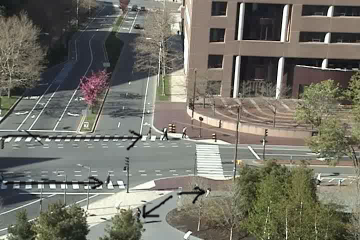}}
  \subfigure[Clusters formed on the frame]{\includegraphics[scale=0.22]{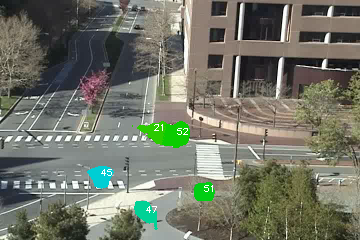}}
  \subfigure[Traces of the five clusters]{\includegraphics[scale=0.22]{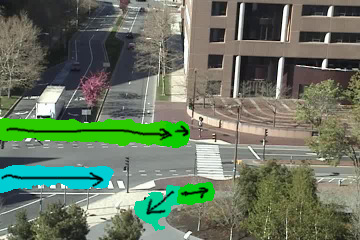}}
  \caption{MIT pedestrian movement detection and tracking. It can be observed that cluster label 21 corresponds to a group of people crossing the zebra line. The Trace for 52 is formed from 21 when the group split into two clusters.}
  \label{Fig:MIT_HUMAN}
\end{figure}

\begin{figure}[h]
  \centering
  \subfigure[Frame marked with motion trace of pedestrians]{\includegraphics[scale=0.25]{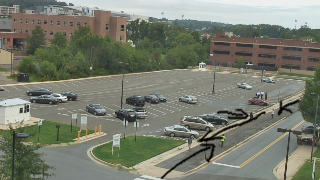}}
  \subfigure[Clusters formed on the frame]{\includegraphics[scale=0.25]{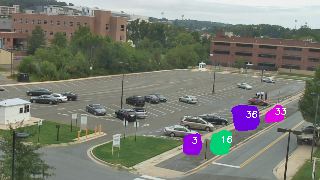}}
  \subfigure[Traces of the 4 pedestrians]{\includegraphics[scale=0.25]{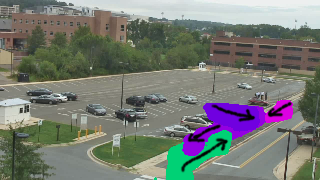}}
  \caption{Detection and tracking of 4 pedestrians. It can be observed that the model was able to discriminate between object 3 and 16 even in close proximity while crossing each other and the traces are maintained throughout the entire episode.}
  \label{Fig:VIRAT_HUMAN}
\end{figure}

We have also tested our model on human motion analysis. Our proposed framework can track individuals when they are spatially apart in video frames. Results of such experiments on MIT and UCF crowd datasets are presented in Fig.~\ref{Fig:VIRAT_HUMAN} and Fig.~\ref{Fig:MIT_HUMAN}. As we consider spatial closeness in the form of distance function, we are able to cluster individuals moving on road with reasonably good accuracy. This indicates that the model can be employed on other objects as long as the objects are not changing their shapes in temporal domain significantly between successive frames. The model can be used to detect abnormal movements of pedestrians while crossing roads via zebra lines, or they are coming on the way of vehicular traffic. Fig.~\ref{Fig:MIT_HUMAN} shows two pedestrians walking on a designated lane. Even though they are in close proximity toward the end of their paths, our model was successful in tracking them correctly. 

\subsection{Overall Comparison}
Our model does clustering and tracking together. KLT tracker~\cite{tomasi1991detection} is an algorithm for feature tracking, even though specific methods can be used for object tracking. Our model can cluster the number of moving objects non-parametrically unlike K-means~\cite{KMeansJin2010} that needs to specify the number of objects and applies a best fit strategy for the objects. Mean-shift clustering~\cite{MeanshiftYizongCheng} which is strictly non-parametric is well suited for clustering. However, it takes multiple iterations to converge. Thus tracking using mean-shift needs special association algorithm to correlate the objects between frames.

\begin{figure*}[h]
  \centering
  \subfigure[KLT]{\includegraphics[scale=0.078]{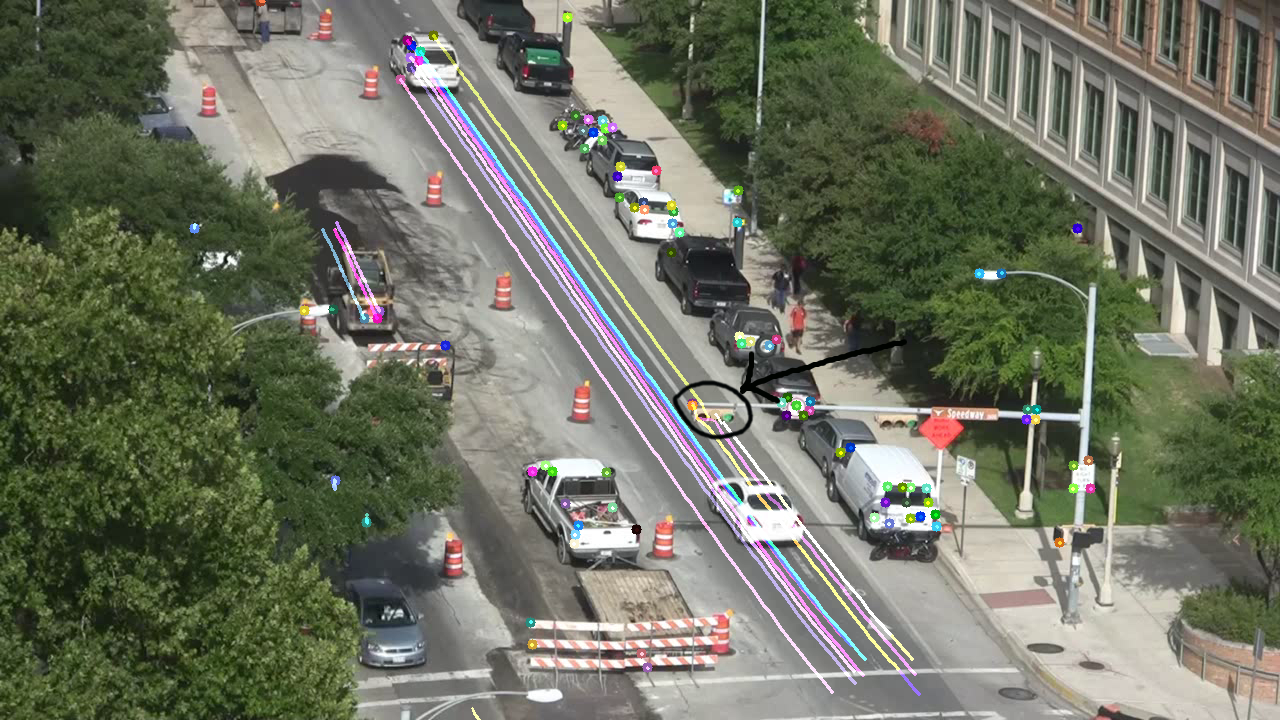}}
  \subfigure[Meanshift]{\includegraphics[scale=0.078]{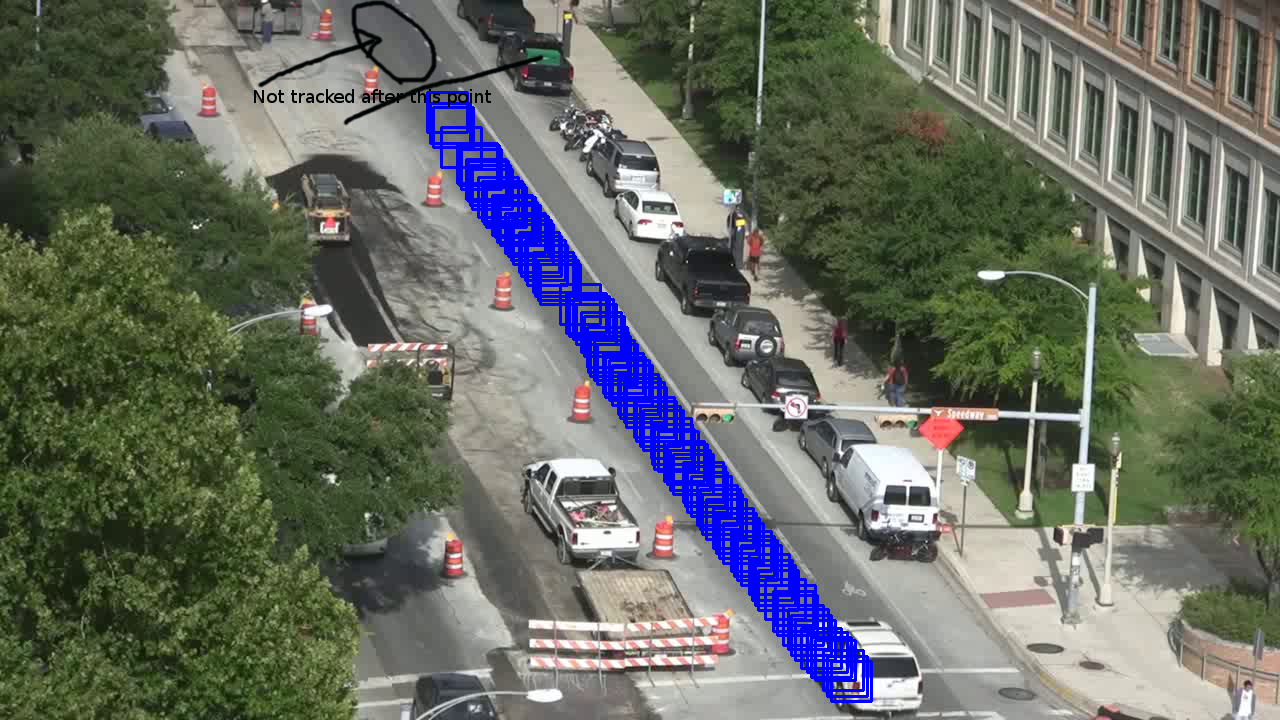}}  
  \subfigure[TLD]{\includegraphics[scale=0.078]{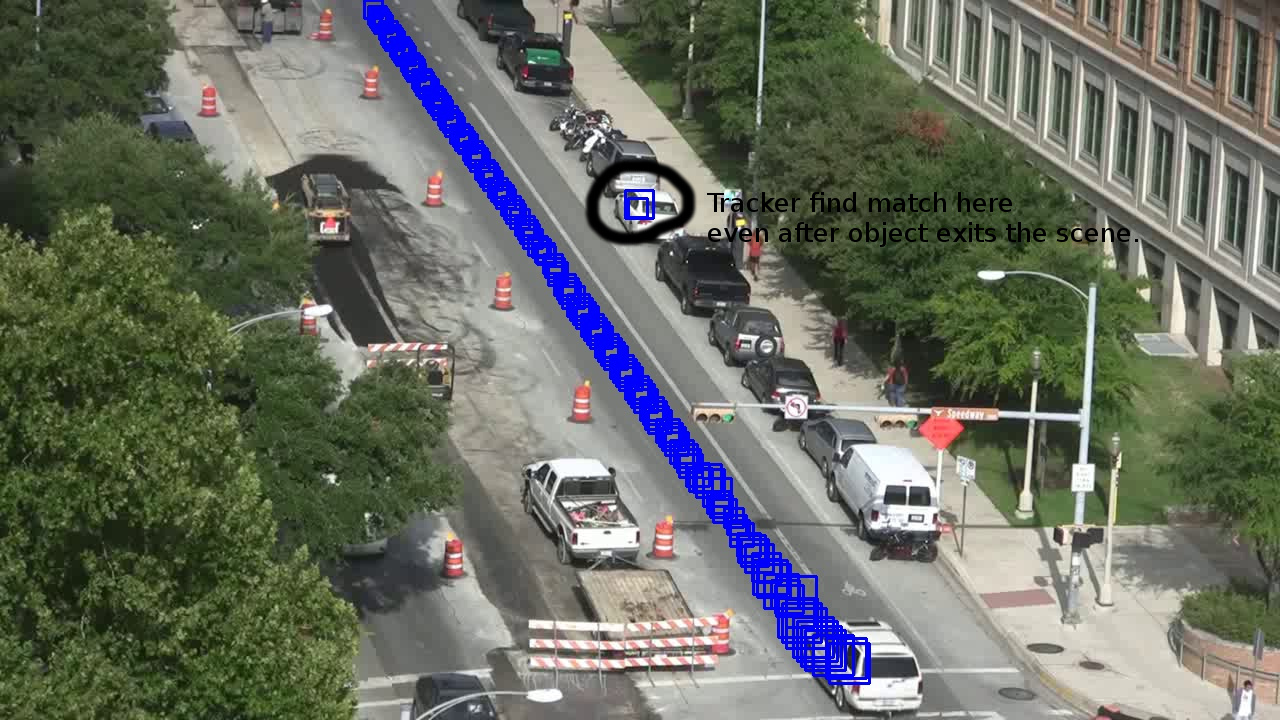}}  
  \subfigure[KCF]{\includegraphics[scale=0.078]{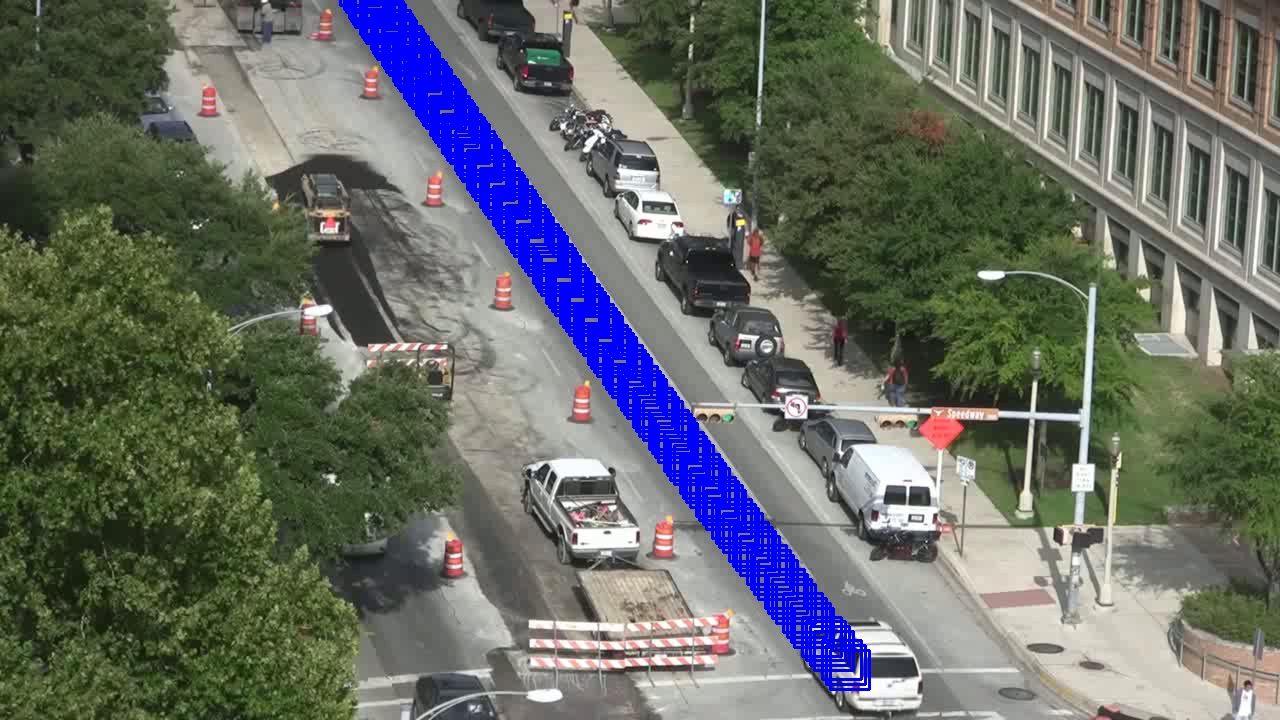}}       
  \subfigure[TUIC]{\includegraphics[scale=0.312]{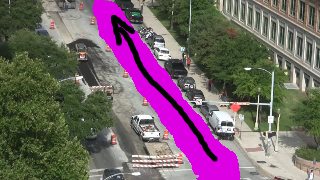}}
  \caption{ (a) KLT tracker being a feature tracker gets affected by occlusion. Three feature points get stopped at the traffic light. A new point is wrongly added to the car. (b) As Mean-shift tracking  does not include tracking features, some heuristic needs to be used for associating objects between consecutive frames. After choosing a region in the video for tracking, it has been found that the tracking stops towards the boundary. (c) TLD finds matching regions even after object exits the scene. (d) KCF tracks the object throughout its presence in the scene. (e) Our proposed TUIC model also tracks the object throughout its life time as can be verified from the figure without needing any initialization.}
  \label{Fig:Track}
\end{figure*}

However, our model does not produce a crisp boundary of the object, rather gives the area of object motion. In normal circumstances, even a human observer looking at a traffic scene may not be always looking at car details like the model or number plate. Rather the human observer may be interested in such details whenever something unusual happens. Since our proposed model provides the patch of the object, with adequate post-processing, finer details of the objects can be obtained. In terms of computational overhead, there is no algorithm that runs in lesser time than $\Theta (nk)$. Hence in real-time applications, our algorithm is best suited. The model is  strictly hierarchical in a sense that we are building pixels to clusters to trajectories. These trajectories can further be used for finding most frequently used patterns as done using a complex VLTAMM model proposed in~\cite{2}. Our model can find the frequently used paths incrementally without any need of the whole video~\cite{2}. A summary of comparisons with state-of-the-art algorithms is presented in Table \ref{tab:Comparison}, where $i$ denotes the number of iterations specified.
\newcolumntype{R}[2]{%
    >{\adjustbox{angle=#1,lap=\width-(#2)}\bgroup}%
    l%
    <{\egroup}%
}
\newcommand*\rot{\multicolumn{1}{R{50}{1em}}}
\begin {table}[H]
\caption {COMPARISON WITH STATE-OF-THE-ART} \label{tab:Comparison} 
\begin{tabular}{r|ccccccccc}
&
\rot{Unsupervised} &
\rot{Non-parametric} &
\rot{Variable sized/shaped clustering} &
\rot{Temporal clustering} &
\rot{Embedded Tracking} &
\rot{Online-Classification} &
\rot{Abnormality Detection} &
\rot{Incremental} &
\rot{Complexity} 

    \\ \hline
TUIC    &  $\checkmark$    &  $\checkmark$    &  $\times$    &  $\checkmark$    &  $\checkmark$    &  $\checkmark$     & $\checkmark$      &  $\checkmark$    &  O($kn$)\\ 
Mean-Shift     &  $\checkmark$    &  $\checkmark$    &  $\times$    &  $\times$ &  $\times$    & $\times$     & $\times$     &  $\times$    &  O($kn^2$)\\ 
K-means      &  $\checkmark$    &  $\times$     &  $\times$    &  $\times$    &  $\times$    & $\times$     & $\times$     &  $\times$    &  O($ikn$)\\ 
DBSCAN     &  $\checkmark$    &  $\checkmark$    &  $\checkmark$    &  $\times$    &  $\times$    & $\times$ & $\times$  &  $\times$   &  O($nlogn$)\\  \hline
\end{tabular}
\end {table}
Tracking performance has been compared against KLT, mean-shift, TLD~\cite{TLD} and KCF~\cite{KCF} algorithms and the results are shown in Fig.~\ref{Fig:Track}. It has been observed that, mean-shift looses tracks when bigger region-of-interest (ROI) is given for tracking. Even it looses the track when the objects reach near the boundary. However, KLT is able to track individual points accurately as it tracks the feature points till the end. Though TLD and KCF algorithms are able to track the objects, however, they require ROI initialization for successful tracking. On the other hand, our algorithm automatically detects the ROI in each frame and temporal association is obtained. TLD tries to match the objects even after they exit the scene. Our model also tracks the object as long as they remain within the scene and are not fully occluded.

\subsection{Analysis of Computational Complexity} 
Our algorithm initializes all observations ($o_i$) to the background cluster $c_0$. During clustering, each observation is unassigned exactly once and checked against each of the $k$ alive clusters to find the probability association with one of the clusters. Thus, if a frame has $n$ pixels with motion constituting $k$ objects, the worst case complexity of the clustering is $\Theta(nk)$. Under normal circumstances, the $k$ will be much smaller than $n$, hence the complexity can be approximated by $\Theta(n)$. It has been observed that, on videos with frame dimension 120 x 213 (approximately $25000$ or more observations) and with number of clusters between $3$ to $20$, our algorithm takes approximately $21$-$27$ ms per frame, when tested on a machine with i5 processor having 4GB or memory. Therefore, we can process all 25 frames within one second for a video recorded at 25 fps. Thus, the proposed model can be used for real-time applications.

\subsection{Limitations}
Even though post-processing can be done to join the connected clusters to handle objects of varying sizes, the method without post-processing can be primarily used for  tracking objects of similar size. The model can track objects with partial occlusion,  the model is not designed to handle full occlusions. However, this provides the base for modeling the lifetime of a cluster. Another issue is, we have used Euclidean distance as the measure to find the distance of an observation from a cluster center. Many of the real life objects like vehicles follow elliptical shape distribution. Therefore, taking maximum Euclidean distance may group pixels belonging to different vehicles moving in similar direction to a single cluster. However, if we take smaller $\beta$ corresponding to the width of the smallest vehicle, the issue can be solved as the vehicles can be separated using connected component analysis.  

\subsection{Summary}
It is found that the algorithm, when applied on vehicle data set, was able to label the clusters as well as track the clusters across different frames with single iteration of Gibbs sampling. In addition, experiments have been carried out on other objects (Human) and it has been found that the proposed model is able to cluster the objects and track them successfully, thus forms a perfect model for traffic analysis.  Performance of the clustering algorithm has been tested and the results reveal that the model can be used for real-time object tracking. The method described for finding the concentration parameter gives a different perspective of the well-known Dirichlet Process~\cite{Farnebäck2003}. 

\section{Conclusion}
\label{sec:conclusion}
This paper introduces an object model from DPMM with a new perspective using a distance measure. The model is temporally extended to consider spatial as well as temporal aspects of moving objects. An incremental approach has been used to build objects from pixels in a hierarchical way without needing to have a prior or the number of clusters. The model has been validated on a wide range of video datasets. The proposed model is able to cluster pixels corresponding to objects and thus can be used to track objects as long as they remain in motion even with partial occlusion. Our model can be applied to videos for building real-time traffic analysis framework as it can learn the segments hierarchically and non-parametrically.

We foresee room for improvement at different levels. Firstly, our model assumes the videos to be shot from the top view. However, most of the videos are not shot accordingly. Secondly, we cannot assume the objects to have fixed dimensions in real life scenario. In such cases, the concentration parameter needs to be automatically learned for each object. Lastly, since we have used optical flow for clustering, further processing may be needed for better object identification as optical flow does not give crisp boundaries of objects. The method in turn can be extended to hierarchically find out most frequently traveled segments of a road.
\bibliographystyle{plain}

\begin{IEEEbiography}[{\includegraphics[width=1in,height=1.25in,clip,keepaspectratio]{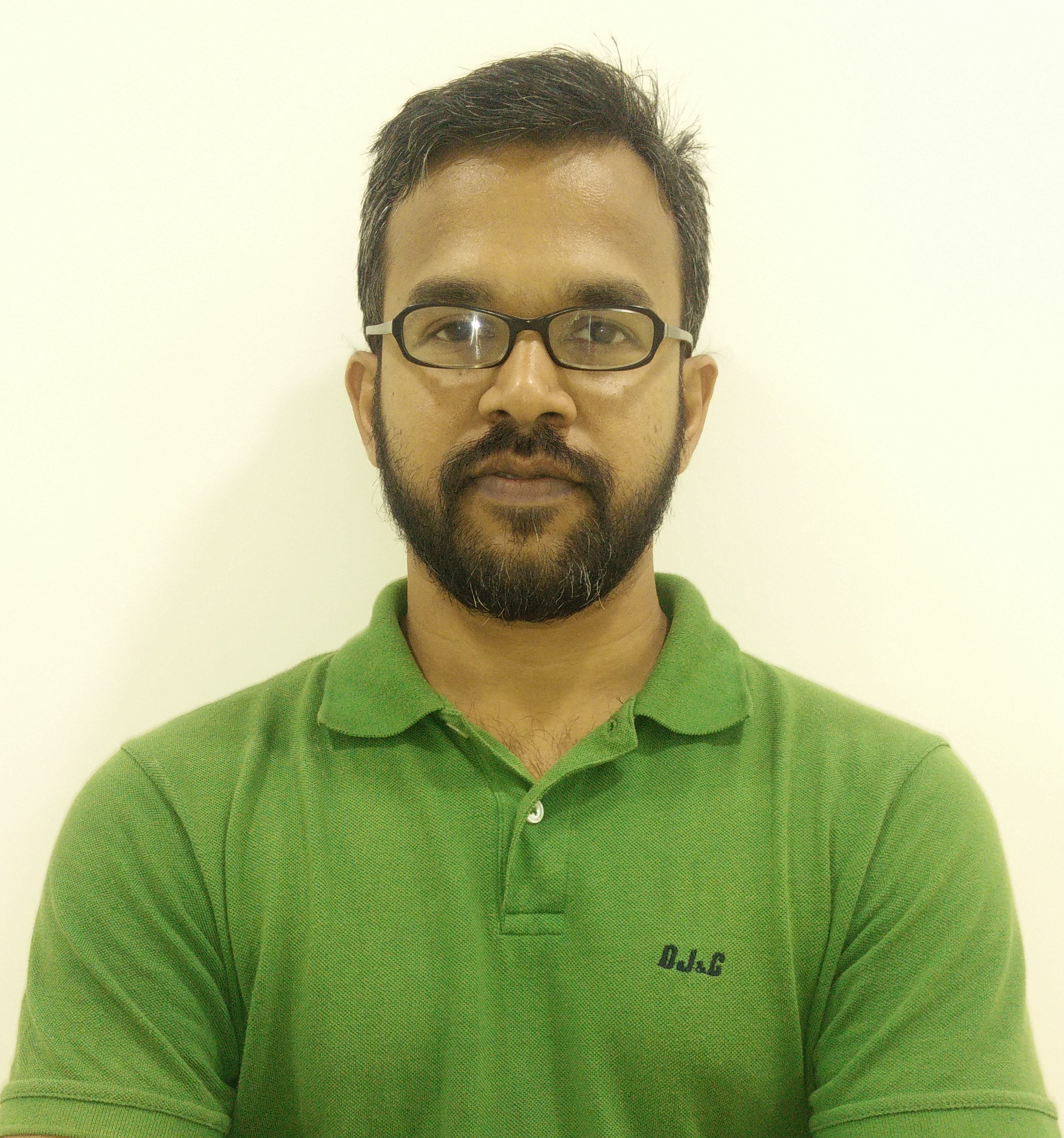}}]{Kelathodi Kumaran Santhosh} is a research scholar in the School of Electrical Sciences, IIT Bhubaneswar, India. He joined a Ph.D. program for resuming his research work that can help humanity. His interests are in the development of vision based applications that can replace human factor. He is a member of IEEE. Prior to joining IIT Bhubaneswar, he worked for Huawei Technologies India Pvt. Ltd. for 10 years (2005-2015) and in Defence Research Development Organization (DRDO) as a Scientist for around 2 years (2003-2004). During his tenure with Huawei, he has worked in many signalling protocols such as Diameter, Radius, SIP etc. in the role of a developer, technical leader, project manager and also served the product lines HSS, CSCF etc. in Huawei China as a support engineer for closer to 1.5 years. In DRDO, he worked in the field of object tracking algorithms based on the data received from radars. More information on Santhosh can be found at https://sites.google.com/site/santhoshkelathodi.
\end{IEEEbiography}
\vspace{-1cm}
\begin{IEEEbiography}[{\includegraphics[width=1in,height=1.25in,clip,keepaspectratio]{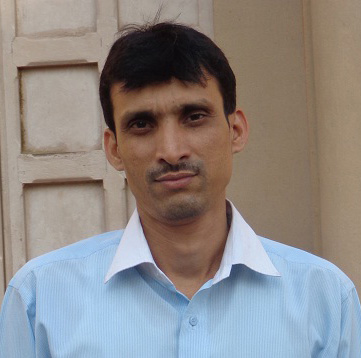}}]{Dr. Debi Prosad Dogra} is an Assistant Professor in the School of Electrical Sciences, IIT Bhubaneswar, India. He received his M.Tech degree from IIT Kanpur in 2003 after completing his B.Tech. (2001) from HIT Haldia, India. After finishing his masters, he joined Haldia Institute of Technology as a faculty members in the Department of Computer Sc. \& Engineering (2003-2006). He has worked with ETRI, South Korea during 2006-2007 as a researcher. Dr. Dogra has published more than 45 international journal and conference papers in the areas of computer vision, image segmentation, and healthcare analysis.  He is a member of IEEE. More information on Dr. Dogra can be found at \url{http://www.iitbbs.ac.in/profile.php/dpdogra}.
\end{IEEEbiography}
\vspace{-1cm}
\begin{IEEEbiography}[{\includegraphics[width=1in,height=1.25in,clip,keepaspectratio]{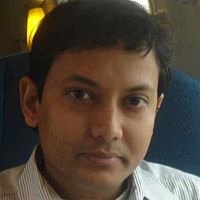}}]{Dr. Partha Pratim Roy} has obtained his M.S. and Ph. D. degrees in the year of 2006 and 2010, respectively at Autonomous University of Barcelona, Spainis. Presently he is an Assistant Professor in the Department of Computer Science and Engineering, IIT Roorkee, India in 2014. Prior to joining, IIT Roorkee, Dr. Roy was with Advanced Technology Group, Samsung Research Institute Noida, India during 2013-2014. Dr. Roy was with Synchromedia Lab, Canada in 2013 and RFAI Lab, France in 2012 as postdoctoral research fellow. His research interests are Pattern Recognition, Multilingual Text Recognition, Biometrics, Computer Vision, Image Segmentation, Machine Learning, and Sequence Classification. He has published more than 65 papers in international journals and conferences. 
\end{IEEEbiography}

\end{document}